\pdfoutput=1

\documentclass[11pt]{article}

\usepackage[final]{acl}

\usepackage{times}
\usepackage{latexsym}

\usepackage[T1]{fontenc}

\usepackage[utf8]{inputenc}

\usepackage{microtype}

\usepackage{inconsolata}

\usepackage{graphicx}

\usepackage{amsmath}
\usepackage{amssymb}
\usepackage{bbold}
\usepackage[noend]{algpseudocode}
\usepackage{algorithmicx,algorithm}
\usepackage{float}
\usepackage{booktabs}
\usepackage{multirow}
\usepackage{makecell}
\usepackage{ulem}
\usepackage{color}
\usepackage{stackengine}
\usepackage{scalerel}

\title{ALiiCE: Evaluating Positional Fine-grained Citation Generation}

\author{Yilong Xu$^{1,2,3}$\quad Jinhua Gao$^{1,2}$\thanks{Corresponding author.} \quad Xiaoming Yu$^{1,2}$\quad Baolong Bi$^{1,2,3}$ \\ \textbf{Huawei Shen}$^{1,2,3}$\quad \textbf{Xueqi Cheng}$^{1,2,3}$\quad \\
$^1$State Key Lab of AI Safety, Institute of Computing Technology, CAS \\
$^2$Key Lab of AI Safety, Chinese Academy of Sciences \\
$^3$University of Chinese Academy of Sciences \\
\texttt{\small{\{xuyilong23s, gaojinhua, yuxiaoming, bibaolong23z, shenhuawei, cxq\}@ict.ac.cn}}
}

\begin{document}
\maketitle
\begin{abstract}
Large Language Model (LLM) can enhance its credibility and verifiability by generating text with citations. However, existing research on citation generation is predominantly limited to sentence-level statements, neglecting the significance of positional fine-grained citations that can appear anywhere within sentences. To facilitate further exploration of the positional fine-grained citation generation, we propose ALiiCE, the first automatic evaluation framework for this task. Our method employs a dependency tree based approach to parse the sentence-level claim into atomic claims. Then ALiiCE evaluates citation quality using three metrics, including positional fine-grained citation recall, precision, and coefficient of variation of citation positions. We evaluate the positional fine-grained citation generation performance of several LLMs on long-form QA datasets. Our experiments and analyses demonstrate the effectiveness and reasonableness of ALiiCE. We offer our insights into the current advancements and future directions for the positional fine-grained citation generation task.\footnote{Our code is available at \url{https://github.com/ylXuu/ALiiCE}.}
\end{abstract}

\section{Introduction}
\label{sec:introduction}

Large Language Models \citep[LLMs;][]{llm} can improve performance in several NLP tasks by incorporating external knowledge ~\citep{rag}. In order to improve LLMs' credibility, ~\citet{gao-etal-2023-enabling, liu-etal-2023-evaluating} propose a new paradigm for long-form QA, in which LLMs are required to provide citations to the retrieved passages for the statements they generate. Since then, many studies ~\citep{ye2024effective, huang2024training, slobodkin2024attribute} have focused on how to enhance LLMs' citation generation capabilities. 

\begin{figure}[t!]
  \includegraphics[width=\columnwidth]{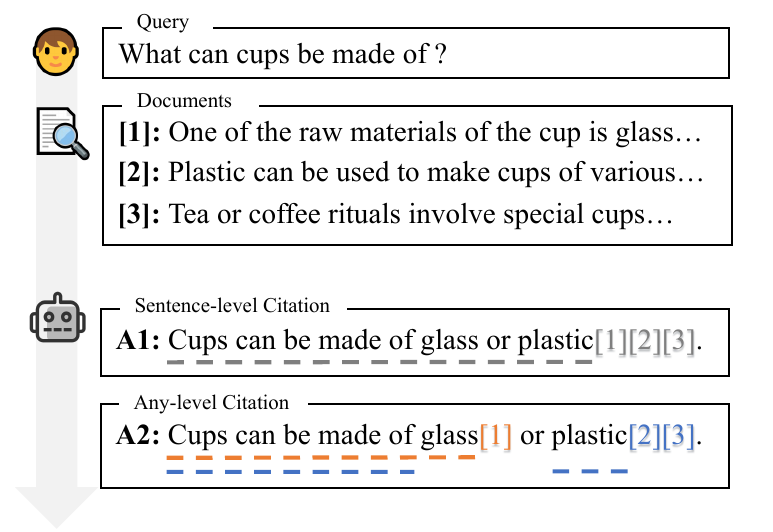}
  \caption{"Sentence-level" vs. "Any-level" in the task of citation text generation. The text with grey underline corresponds to the claim in A1 cited by "[1][2][3]". The texts of orange and blue underlines correspond to the claims in A2 cited by "[1]" and "[2][3]", respectively.}
  \label{fig:intro}
\end{figure}

However, existing research on citation generation is predominantly limited to sentence-level statements. ~\citet{malaviya2024expertqa} suggest that a sentence might not be the smallest unit capable of representing an atomic claim, potentially leading to inaccurate evaluations. As shown in Figure \ref{fig:intro}, response A1 actually contains two different claims, but the sentence-level citation treats the entire sentence as one claim. Additionally, ~\citet{liu-etal-2023-evaluating} highlight that the generated text scope of a single in-line citation is often ambiguous. Citations of A1 in Figure \ref{fig:intro} is ambiguous, because the citation marks at the end of A1 do not clearly indicate whether they support both claims or only the last claim.

In fact, in many long-form contexts, particularly in professional fields such as academic writing ~\citep{funkquist2023citebench}, citation marks often appear in the middle of a sentence rather than always at the end, as response A2 illustrated in Figure \ref{fig:intro}. Compared with sentence-level citation, the advantages of this fine-grained generation are: 
1) clearer indication of the text scope associated with each citation mark, and 2) better user-friendliness, allowing users to locate more specific content to check. We refer to this improved generation task as Positional Fine-grained Citation Text Generation.

Despite the importance of this task, an effective evaluation method has yet to be developed. Some studies directly apply sentence-level metrics to fine-grained citations \citep{huang2024learningfinegrainedgroundedcitations}, but this can affect the accuracy of evaluation. First, sentence-level metrics simply merge evidences of multiple atomic claims ~\citep{gao-etal-2023-enabling}. When using Natural Language Inference \citep[NLI;][]{nli} to judge entailment, merged evidences can easily result in the issue of excessively long NLI contexts. Second, if there is an overlap between evidence of different atomic claims, sentence-level judgments can also become unreasonable, for correct citations might be mistakenly excluded. Thus, it is essential to design an evaluation method specifically tailored for positional fine-grained citations.

We propose a new evaluation method, \textbf{ALiiCE}, \textbf{A}utomatic \textbf{L}LM's Pos\textbf{i}tional F\textbf{i}ne-grained \textbf{C}itation \textbf{E}valuation. Our method first employs a Dependency Tree based approach to parse atomic claims of each citation in the response. For instance, the two claims of A2 in Figure \ref{fig:intro} are parsed as "Cups can be made of glass" and "Cups can be made of plastic". Further, our method incorporates three metrics for evaluating citation quality, including citation recall and precision at the level of atomic claims, as well as the Coefficient of Variation of Citation Positions (CVCP), which measures the dispersion of citation positions within a sentence.

In our experiment, we employ two long-form QA datasets, ASQA ~\citep{stelmakh-etal-2022-asqa} and ELI5 ~\citep{fan-etal-2019-eli5} to evaluate outputs of LLMs including GPT-3.5, GPT-4 and LLaMA-3-8B. We observe that existing LLMs generate a limited number of positional fine-grained citations. We compare the citation quality of LLMs' outputs in sentence-level metrics with ALiiCE to demonstrate the necessity of evaluation method for positional fine-grained citations. We also conduct error analyses to assess the impact of parsing errors. Additionally, we conduct human evaluation to verify the consistency between ALiiCE and human judgment.



To summarize, our main contributions include:
\begin{itemize}
    \item We propose the first dedicated evaluation method for positional fine-grained citation generation and we prove its effectiveness through experiments;
    \item We evaluate the performance of several existing LLMs on positional fine-grained citation generation in long-form QA datasets;
    \item We offer our insights on the study of positional fine-grained citation generation: 1) open-source LLMs show great progress in citation generation, substantially narrowing the gap with closed-source LLMs; 2) feedback from human evaluation suggests that existing citation evaluation methods still overlook citation utility, which is a crucial aspect of assessing citation quality.
\end{itemize}
We hope that our work can inspire more research into positional fine-grained citation text generation.

\begin{figure*}[htbp]
  \includegraphics[width=\linewidth,scale=1.00]{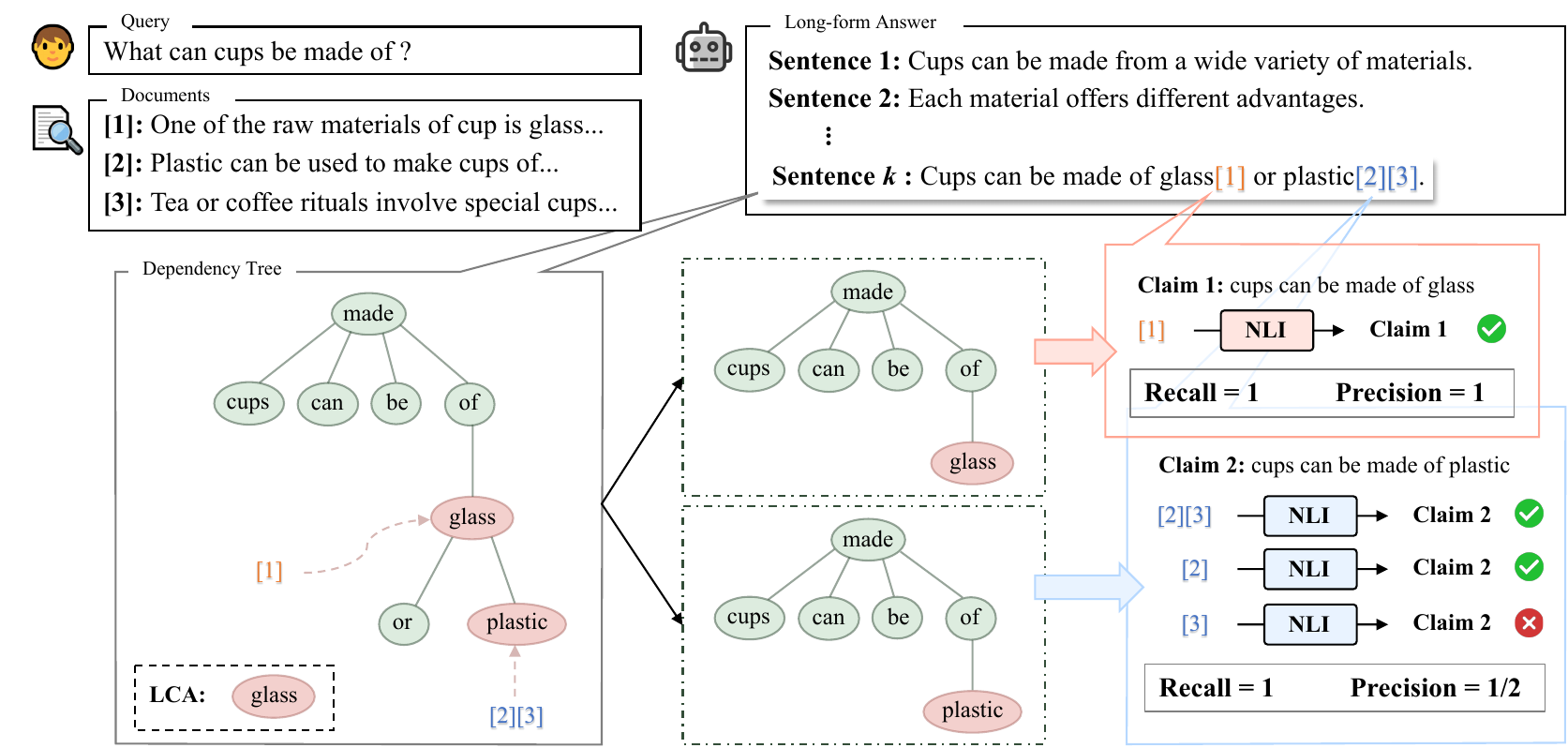}
  \caption {An example of ALiiCE evaluation framework on positional fine-grained citation generation. Given a query and related documents, the LLM generates a long-form answer. For sentence $i$ in answer, the parsing pipeline involves constructing the dependency tree, identifying the LCA node to obtain the modified tree of each claim, and converting modified trees into texts. Finally, we calculate the citation recall and precision for each claim.}
  \label{fig:framework}
\end{figure*}

\section{Background \& Task Definition}

In this section, we briefly introduce the background of our research and provide a definition of positional fine-grained citation generation.

\subsection{Citation Generation in Long-form QA}

Unlike short-form QA, which typically provides binary, entity-level, or short sentence answers, long-form QA generates detailed and comprehensive responses including explanations, context, and additional relevant information \citep{krishna-etal-2021-hurdles}. Citation generation involves producing citation marks (namely, passage IDs) while generating text, indicating the source passages on which the text is based ~\citep{funkquist2023citebench, huang2024citationkeybuildingresponsible}. In our work, we focus on positional fine-grained citation generation for long-form QA. Unlike traditional task, it allow citation marks to appear at any position within the sentence.

\subsection{Task Definition}

Formally, given a query $q$ and a set $\mathcal{D}$ of retrieved passages based on $q$, the generator $\mathcal{M}$ is required to generate a long-form response $\mathcal{R}$ containing citations. Specifically, $\mathcal{R}$ is composed of several sentences, with each sentence containing words and in-line citation markers. 
We assume that the $k$-th sentence $s_k$ has a length of $l$ and can be represented as $x_1, x_2, \ldots, x_l$, where $x_i$ represents the $i$-th minimal semantic unit in $s_k$. 

In this paper, the minimal semantic unit can be either a word (including punctuation) or a group of citation marks. If $x_i$ is the latter, we denote it as $\mathcal{C}_i=\left\{ c_{i,1}, c_{i,2}, \ldots \right\}$, where $c_{i,j}$ is a citation mark of a passage in $\mathcal{D}$. And $C_i$ corresponds to an atomic claim parsed from its sentence, marked as $\mathcal{A}_i$. Take A2 in Figure \ref{fig:intro} as an example, "plastic" is a word, and "[2][3]" is a group of citation marks with its atomic claim "Cups can be made of plastic".

\section{ALiiCE: Automatic LLMs' Positional Fine-grained Citation Evaluation}

In this section, we give a detailed description of ALiiCE. First, we introduce how we construct the atomic claim parsing pipeline based on dependency trees. Then, we present three metrics for the evaluation of positional fine-grained citation quality.

\subsection{Dependency Tree}
Dependency trees are hierarchical representations of the grammatical structure of a sentence, showing how words rely on each other ~\citep{culotta-sorensen-2004-dependency}, and is more concise compared with the hierarchical syntax tree based on operators. In a dependency tree, a subtree can represent a phrase or clause that depends on its root, which is highly suitable for atomic claims extraction. Thus in ALiiCE, we employ dependency trees to represent sentences in $\mathcal{R}$ for subsequent parsing stage.

For simplicity, we assume that the nodes in the dependency tree are all words. To extract atomic claim based on the position of the $\mathcal{C}_i$ in the original sentence, we find a matching node in the tree for each $\mathcal{C}_i$, as shown in the lower left part of Figure \ref{fig:framework}. In this paper, we refer to the node with citation marks attached as the citation node.

When handling multiple citations in different positions within a sentence, their respective claims need to be parsed. To parse the claim of $\mathcal{C}_i$, we need to exclude irrelevant content from other claims, as different claims may share identical sentence components. Thus, we consider the Lowest Common Ancestor (LCA), which is the deepest node of two different nodes possessing both of them as descendants in a tree. For two distinct citation nodes, we can modify the dependency tree to obtain atomic claims based on the relative positions with respect to their LCA node (see Section \ref{sec:algorithm}).

\subsection{Parsing Pipeline}
\label{sec:algorithm}

Our parsing pipeline is illustrated by Figure \ref{fig:framework} and simplified pseudo code is shown in Algorithm \ref{algo:algo_1}.

For sentence $s_k$ in response $\mathcal{R}$, we extract groups of citation marks $\left\{ \mathcal{C}_i \right\}$ from difference positions. Then we do text cleaning on $s_k$ to obtain raw sentence ${s_k}'$, involving removing citation marks and other punctuation. ${s_k}'$ is used to construct dependency tree $T$. Next, we match citation node for each $\mathcal{C}_i$. The principle of matching node is to select word closest to $\mathcal{C}_i$ in $s_k$, giving priority to the one before $\mathcal{C}_i$. Then we modify the dependency tree based on the citation nodes.

For each citation node, denoted as node $i$, iterate other citation nodes except node $i$. When iterating to node $j$, we calculate the LCA node of node $i$ and node $j$ in $T$. Then we find the subtrees of LCA node's children containing node $i$ and node $j$, and denote them as $T_i$ and $T_j$, respectively. Next, we discuss in different situations:

\begin{itemize}
    \item If LCA node is node $i$, remove $T_j$ from $T$.
    \vspace{-8pt}
    \item If LCA node is node $j$, replace LCA node's subtree with $T_i$.
    \vspace{-8pt}
    \item If LCA node is another node in $T$, then we compare the relative positions between $T_i$ and $T_j$, according to the word's order in the sentence of subtree's root: If $T_i$ is before $T_j$, then remove $T_j$ from $T$; If $T_i$ is after $T_j$, then replace LCA node's subtree with $T_i$. 
\end{itemize}

After iteration, we obtain a modified dependency tree. We convert words in the modified tree to text following the order in original sentence, getting the claim of citations corresponding to node $i$. We provide additional details and running examples of our algorithm in Appendix \ref{appendix:parsing_algorithm_details} and \ref{appendix:parsing_examples}, respectively.

\newcommand{\MyIndent}[1]{
    \hspace{#1\algorithmicindent}
}

\begin{algorithm}[t]
\caption{ALiiCE's Parsing Algorithm}
\hspace*{0.02in} {\bf Input:}
A sentence $s$ with in-line citation marks\\
\hspace*{0.02in} {\bf Output:}
A list of claim of each group of citation marks
\begin{algorithmic}[1]
\State $L=\phi $
\State ${s}'=\textsc{TextCleaning} \left( s \right)$ 
\State $T = \textsc{DependencyTree} \left( {s}' \right)$ 
\State $nodes = \textsc{MatchCitationNodes} \left( T,s \right)$ 
\For{each $node_i$ in $nodes$}
\State{${T}'=\textsc{DeepCopy} \left(T \right)$}
\For{each $node_j$ in $nodes \setminus \left\{ node_i \right\}$}
\State $node_{lca} = \textsc{LCA} \left( {T}', node_i, node_j \right) $ 
\State $T_i=\textsc{SubTree} \left( node_{lca},node_i \right)$ 
\State $T_j=\textsc{SubTree} \left( node_{lca},node_j \right)$

\State \textbf{if} $node_{lca}=node_i \vee$ \\
\hspace{\algorithmicindent}\hspace{\algorithmicindent}
$\left( node_{lca}\neq node_j \wedge T_i < T_j \right)$ \textbf{then}

\State \hspace{\algorithmicindent} $\textsc{Mask} \left({T}', T_j \right)$
\State \textbf{else}
\State \hspace{\algorithmicindent} $\textsc{Replace} \left( node_{lca}, T_i \right)$ 

\EndFor
\State $r = \textsc{ConvertToText} \left( {T}' \right)$ 
\State $r \longrightarrow  L$
\EndFor
\State \Return $L$
\end{algorithmic}
\label{algo:algo_1}
\end{algorithm}

\subsection{Metrics for Citation Quality}
\label{sec:metrics}

In this section, we display the three metrics for positional fine-grained citation quality in ALiiCE.

\subsubsection{Positional Fine-grained Citation Recall}
\label{sec:positional_fine_grained_citation_recall}

For each $\mathcal{C}_i$ and its corresponding $\mathcal{A}_i$, if the concatenation of passages in $\mathcal{C}_i$ can entail $\mathcal{A}_i$, then the citation recall is 1, otherwise it is 0. The judgement of entailment can be formulated as:

\begin{equation}
\Psi \left(\mathcal{H}, \mathcal{S}\right)=\begin{cases}
1,& \text{if} \; \mathcal{H} \; \text{entails} \; \mathcal{S}\\
0,& \text{else}
\end{cases}
\end{equation}
where $\Psi$ represents a NLI model, and $\mathcal{H}$ and $\mathcal{S}$ represent hypothesis and statement, respectively.

\subsubsection{Positional Fine-grained Citation Precision}
\label{sec:positional_fine_grained_citation_precision}

Following ~\citet{gao-etal-2023-enabling}, we calculate citation precision to evaluate whether every citation is necessary. This metric checks for redundant citations to improve readability and verifiability.

We compute citation precision only when the citation recall of $\mathcal{C}_i$ is 1; otherwise, the citation precision is set to 0. Specifically, for each $c_{i,j}$ in $\mathcal{C}_i$, if $c_{i,j}$ can not entail $\mathcal{A}_i$ alone while the concatenation of passages in $\mathcal{C}_i \setminus c_{i,j}$ can, it is indicated that $c_{i,j}$ is a redundant citation and the precision score of $c_{i,j}$ is 0, otherwise the precision score of $c_{i,j}$ is 1. Finally we calculate the mean of the precision scores from each $c_{i,j}$ as the precision score of $\mathcal{C}_i$.

\begin{table*}
  \centering
  \small
  \resizebox{\linewidth}{!}{
  \renewcommand\arraystretch{1.1}
  \begin{tabular}{clcccccccc}
    \toprule
    \multirow{3}{*}{\textbf{Dataset}} & \multirow{3}{*}{\textbf{Model ($k$-psg-form)}} & \multicolumn{3}{c}{\textbf{ALiiCE}} & \multirow{3}{*}{\textbf{CVCP}} & \multirow{3}{*}{\textbf{Fluency}} & \multirow{3}{*}{\textbf{Correct.}} & \multirow{3}{*}{\textbf{Length}} \\
    \cmidrule(lr){3-5}
     & & \textbf{Rec.} & \textbf{Prec.} & \textbf{F1.} &  &  &  &  \\
     \midrule
     \multirow{7}{*}{ASQA} & GPT-3.5 (5-psg) & \textbf{78.4} $_{(0.5)}$ & 74.4 $_{(0.4)}$ & 76.3 $_{(-)}$ & 0.10 $_{(-)}$ & 86.1 $_{(2.9)}$ & 51.1 $_{(0.3)}$ & 50.5 $_{(37.3)}$ \\
     & GPT-3.5 (5-psg-summ) & 76.9 $_{(0.4)}$ & 71.6 $_{(0.9)}$ & 74.2 $_{(-)}$ & 0.13 $_{(-)}$ & 75.4 $_{(2.3)}$ & 49.3 $_{(0.3)}$ & 40.2 $_{(33.2)}$ \\
     & GPT-3.5 (5-psg-snip) & 74.4 $_{(0.7)}$ & 69.4 $_{(0.3)}$ & 71.8 $_{(-)}$ & 0.13 $_{(-)}$ & 73.1 $_{(3.7)}$ & 48.0 $_{(0.6)}$ & 36.0 $_{(29.7)}$ \\
     & GPT-3.5 (10-psg) & 77.7 $_{(1.2)}$ & \textbf{75.9} $_{(0.8)}$ & \textbf{76.8} $_{(-)}$ & 0.15 $_{(-)}$ & 84.6 $_{(7.1)}$ & 44.1 $_{(0.3)}$ & 63.4 $_{(52.6)}$ \\
     & GPT-4 (5-psg) & 76.8 $_{(1.2)}$ & 68.2 $_{(1.1)}$ & 72.2 $_{(-)}$ & 0.15 $_{(-)}$ & 52.2 $_{(9.5)}$ & 47.0 $_{(0.4)}$ & 28.1 $_{(20.8)}$ \\
     & LLaMA-3-8B (5-psg) & 64.8 $_{(1.0)}$ & 61.4 $_{(1.4)}$ & 63.1 $_{(-)}$ & 0.44 $_{(-)}$ & 84.2 $_{(5.0)}$ & 50.9 $_{(0.3)}$ & 64.0 $_{(53.1)}$ \\
     & LLaMA-3-8B (10-psg) & 61.8 $_{(1.3)}$ & 62.5 $_{(0.5)}$ & 62.1 $_{(-)}$ & \textbf{0.45} $_{(-)}$ & 88.8 $_{(9.6)}$ & 41.7 $_{(1.4)}$ & 73.2 $_{(64.9)}$ \\
    \midrule
     \multirow{7}{*}{ELI5} & GPT-3.5 (5-psg) & \textbf{61.0} $_{(0.5)}$ & \textbf{58.6} $_{(2.2)}$ & \textbf{59.8} $_{(-)}$ & 0.10 $_{(-)}$ & 21.8 $_{(0.6)}$ & 20.8 $_{(0.3)}$ & 131.7 $_{(46.2)}$ \\
     & GPT-3.5 (5-psg-summ) & 53.9 $_{(2.6)}$ & 52.0 $_{(1.1)}$ & 52.9 $_{(-)}$ & 0.15 $_{(-)}$ & 21.3 $_{(5.1)}$ & 20.8 $_{(1.1)}$ & 111.3 $_{(46.5)}$ \\
     & GPT-3.5 (5-psg-snip) & 53.4 $_{(1.3)}$ & 50.9 $_{(1.1)}$ & 52.1 $_{(-)}$ & 0.13 $_{(-)}$ & 34.9 $_{(7.3)}$ & 20.8 $_{(0.4)}$ & 106.7 $_{(47.9)}$ \\
     & GPT-3.5 (10-psg) & 58.1 $_{(2.4)}$ & 56.8 $_{(2.0)}$ & 57.4 $_{(-)}$ & 0.12 $_{(-)}$ & 18.5 $_{(4.7)}$ & 19.7 $_{(0.7)}$ & 155.9 $_{(57.4)}$ \\
     & GPT-4 (5-psg) & 55.1 $_{(0.5)}$ & 54.0 $_{(3.0)}$ & 54.5 $_{(-)}$ & 0.15 $_{(-)}$ & 20.4 $_{(7.2)}$ & 21.3 $_{(0.9)}$ & 102.2 $_{(59.7)}$ \\
     & LLaMA-3-8B (5-psg) & 45.9 $_{(0.3)}$ & 47.1 $_{(0.7)}$ & 46.5 $_{(-)}$ & 0.53 $_{(-)}$ & 36.2 $_{(1.0)}$ & 20.5 $_{(0.9)}$ & 203.9 $_{(71.4)}$ \\
     & LLaMA-3-8B (10-psg) & 42.8 $_{(0.8)}$ & 44.2 $_{(0.9)}$ & 43.5 $_{(-)}$ & \textbf{0.61} $_{(-)}$ & 32.5 $_{(6.2)}$ & 19.5 $_{(0.7)}$ & 224.2 $_{(77.7)}$ \\
    \bottomrule
  \end{tabular}
  }
  \caption{Results on ASQA and ELI5. The $k$-psg indicates using top-$k$ relevant documents for response generation. Document formats include summary (summ), snippet (snip), and default original text. The correctness refers to the exact match recall for ASQA and ROUGE-L for ELI5. The value in bracket represents the standard deviation.}
  \label{table:main_result}
\end{table*}

\subsubsection{Coefficient of Variation of Citation Positions}

\newcommand\mysqrt[2][0pt]{\stretchrel{\sqrt{}}{\addstackgap%
  [#1]{$\displaystyle\overline{#2}$}}}

Positional fine-grained citation generation allows citation marks to appear in multiple positions within a sentence (e.g., in the middle, at the end). Consequently, to some extent, the dispersion of citation marker positions can reflect the LLMs' ability to generate positional fine-grained citations. For example, in Figure \ref{fig:intro}, A2 has a greater dispersion of citation marker positions than A1. To quantify the degree of dispersion, we propose CVCP (\textbf{C}oefficient of \textbf{V}ariation of \textbf{C}itation \textbf{P}ositions). 

For response $\mathcal{R}$, we first calculate the indices of citation marks' positions for every sentence. For sentence $s_k$, which has a length of $l$ and can be represented as $x_1,\ldots,x_l$, we extract the subscripts corresponding to the citation marks as the indices, denoted by $p_1,\ldots,p_t$, where $t$ is the number of citation marks. We normalize the indices to eliminate the interference of sentence length as follows:

\begin{equation}
    p_i \gets \frac{p_i}{\left | s_k \right | }
\end{equation}

Then we compute standard deviation for $s_k$ as:

\begin{equation}
    \sigma \left ( s_k \right ) = \mysqrt[1.1pt]{\frac{1}{t} \sum^{t}_{{j}=1} \left ( p_j- \mu_k \right )^2 }
\end{equation}

where $\mu_k=\frac{1}{t} \sum^{t}_{j=1} p_j$, which represents the mean of normalized indices. Assuming $s_k$ has $n$ sentences, the CVCP of $\mathcal{R}$ is as follows:

\begin{equation}
    CV_{CP}\left( \mathcal{R} \right) = \frac{1}{n} \sum^n_{k=1} \frac{\sigma \left( s_k \right)}{\mu_k}
\end{equation}

When the positions of the citation markers in the sentence are more dispersed, the CVCP can be higher. Conversely, if all citation markers appearing at the end of the sentence, the CVCP can be very low (i.e., 0). Thus, CVCP encourages LLMs to generate more positional fine-grained citations.

\section{Experimental Setup}
In this section, we describe the datasets and implementation details of our experiments. Additional details are provided in Appendix \ref{appendix:experimental_setup_details}.

\paragraph{Datasets} We utilize two popular datasets for the task of long-form QA, including: 1) \textbf{ASQA}, which is an open-domain long-form QA dataset for ambiguous factoid queries, collected from AmbigQA ~\citep{min-etal-2020-ambigqa}; 2) \textbf{ELI5}, which is a dataset for complex QA with paragraph-length responses, collected from subreddit "Explain Like I’m Five". The queries of these two datasets are well suited for retrieval-augmented generation, thus more conducive for evaluating fine-grained citation generation. Following \citet{gao-etal-2023-enabling}, we use the Generalizable T5-based dense Retriever ~\citep[GTR;][]{gtr} to retrieve relevant passages for queries from Wikipedia corpus snapshot dated 2018-12-20.

\paragraph{Implementation} We utilize SpaCy\footnote{\url{https://spacy.io/}} to construct dependency trees for sentences, which is a useful and efficient python toolkit for many NLP tasks. We use TRUE\footnote{\url{https://huggingface.co/google/t5_xxl_true_nli_mixture}}, a fine-tuned T5-11B ~\citep{t5} model as the NLI model for the judgement of entailment in citation quality. 

\newcommand{\hytt}[1]{\texttt{\hyphenchar \font=\defaulthyphenchar#1}}

\paragraph{Models} For closed-source LLMs, we evaluate \hytt{gpt-4-turbo-2024-04-09} and \hytt{gpt-3.5-turbo-0125} ~\citep{chatgpt, openai2024gpt4}. For open-source LLMs, we evaluate \hytt{LLaMA-3-8B} ~\citep{llama3modelcard}. In addition, we incorporate variables such as the number of retrieved passages and the passage form used in generation (truncated original text, summary, or snippet) into the model setting. The prompts are provided in Appendix \ref{appendix:promtps}.

\paragraph{Evaluation Metrics} In addition to the three metrics of citation quality introduced at Section \ref{sec:metrics}, we utilize three common metrics in long-form QA, including: 1) \textbf{correctness}, which checks whether $\mathcal{R}$ answers the query $q$ accurately; 2) \textbf{fluency}, which evaluates whether $\mathcal{R}$ is coherent; and 3) \textbf{length}, which is the average length of $\mathcal{R}$. Regarding correctness, for ASQA, we follow \citet{stelmakh-etal-2022-asqa} to calculate exact match recall by checking whether ground truths are exact substrings of $\mathcal{R}$; for ELI5, we follow \citet{fan-etal-2019-eli5} to use the F1 score of ROUGE-L. We quantify the fluency by MAUVE \citep{mauve}. For comparison, we also use ALCE \citep{gao-etal-2023-enabling} as the sentence-level evaluation to assess citation quality.

\begin{figure}[t!]
  \includegraphics[width=\linewidth]{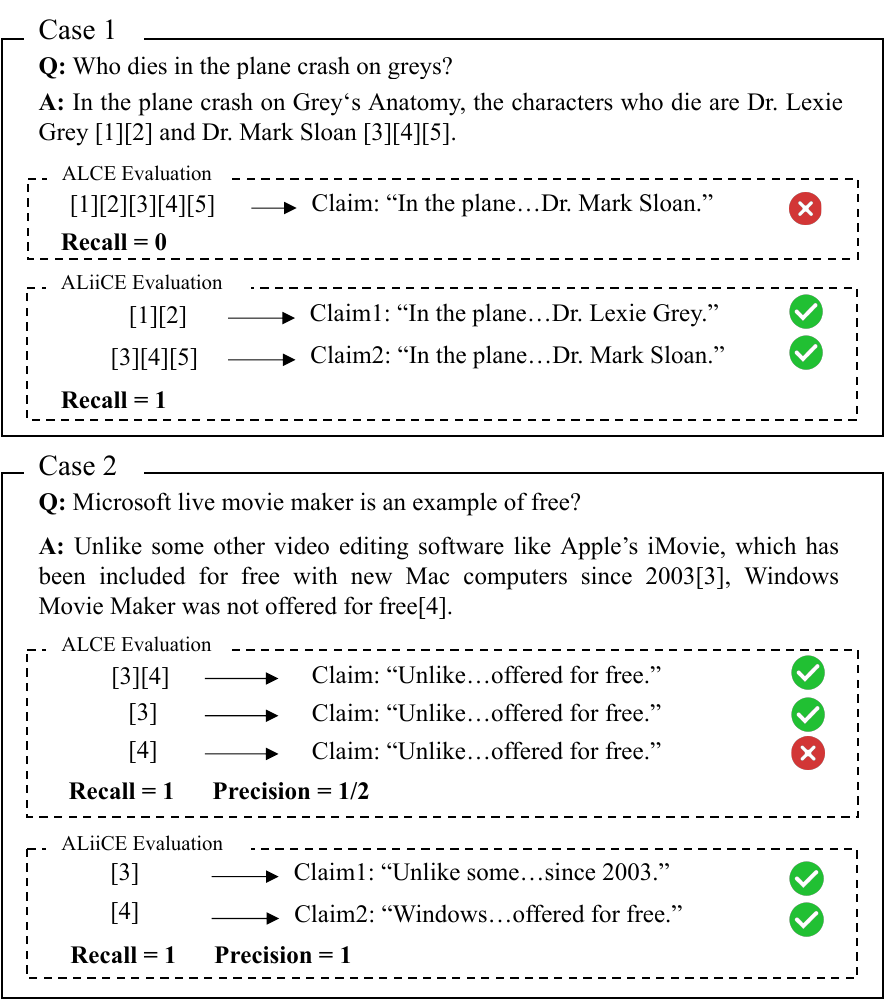}
  \caption {Evaluation process of citation quality by ALCE and ALiiCE on two examples from ASQA. The answers are generated by GPT-3.5 (5-psg). 
  }
  \label{fig:cases}
\end{figure}

\section{Main Results}

In this section, we present our key observations on the experiment results, and then provide our case study to prove the necessity of developing method for positional fine-grained citation evaluation.

\subsection{Overall Performances}
\label{sec:overall_performances}

The result of our experiment is presented in Table \ref{table:main_result}. We obtain some key observations as follows:

\paragraph{Citation quality} In ASQA, GPT-3.5 (10-psg) achieves the best performance in citation recall and precision, while in ELI5, the top performer is GPT-3.5 (5-psg). Overall, these two models exhibit outstanding performance of citation quality across both datasets. Moreover, simpler passage formats, such as summary and snippet, do not yield performance improvements. Through CVCP, we observe that most models generate a limited number of positional fine-grained outputs. LLaMA-3-8B is able to generate more fine-grained samples than GPT-3.5 and GPT-4, among which LLaMA-3-8B (10-psg) achieves the highest CVCP in both datasets.

\paragraph{Other metrics} The difference in fluency between the models is not significant; however, it is evident that the model outputs for ELI5 are much less fluent compared to ASQA. This discrepancy is also observed in terms of correctness. Regarding length, the model outputs for ELI5 are substantially longer than those for ASQA, and the output length of LLaMA-3 is longer than that of GPT-3.5 and GPT-4. As for the difference between the two datasets, we believe that this may be due to the more difficult queries and the more complex knowledge contained in the passages in ELI5.

\begin{table}[t] 
  \centering
  \tiny
  \resizebox{\linewidth}{!}{
  \begin{tabular}{cccc}
    \toprule
     \textbf{Dataset} & \textbf{Num of Claims} & \textbf{Num of same NLI} \\
     \midrule
     ASQA & 1935 & 1930 \\
     ELI5 & 3923 & 3891 \\
    \bottomrule
  \end{tabular}
  }
  \caption{Results on parsing error analyses. The second column is the total number of claims. The last column is the number of claims with consistent NLI results before and after refinement on the claims. }
  \label{table:error_analyses}
\end{table}

\subsection{Case Study}

In this section, we compare the evaluations of ALCE and ALiiCE on two instances from ASQA, and analyze the shortcomings and insufficiency of sentence-level metrics on positional fine-grained citation evaluation. Our objective is to demonstrate the necessity of designing a dedicated citation evaluation method with atomic claim parsing.

\paragraph{Long-context issue} Sentence-level evaluation can result in inaccuracies when dealing with long-context NLI. For instance, in Case 1 depicted in Figure \ref{fig:cases}, when assessing citation recall, the concatenated passages exceed the context length of NLI model, potentially leading to incorrect inference results due to distracted attention or truncation of evidences. In ALiiCE, evidences are dispersed by parsing atomic claims, reducing the likelihood of exceeding context limits. 

\paragraph{Citation precision issue} If there is an overlap between different evidences, it is potential for the NLI model to misjudge multiple atomic claims simultaneously. Taking the Case 2 in Figure \ref{fig:cases} as an example, citation "[3]" contains evidences supporting both atomic claim 1 and 2. According to ALCE's citation precision, citation "[3]" alone can support the entire sentence-level claim, whereas citation "[4]" cannot, as it only supports atomic claim 2. Consequently, citation "[4]" is considered redundant, despite being or even though it is actually a reasonable citation. In ALiiCE, we evaluate based on atomic claims, ensuring that the assessment is not influenced by evidences from other claims.

\begin{table}\small
  \centering
  \resizebox{\linewidth}{!}{
  \renewcommand\arraystretch{1.3}
  \begin{tabular}{lcccc}
    \toprule
    \makecell[l]{\multirow{3}{*}{\textbf{Model ($k$-form)}}}  & \multicolumn{2}{c}{\textbf{ALiiCE}} & \multicolumn{2}{c}{\textbf{ALCE}} \\
    \cmidrule(lr){2-3} \cmidrule(lr){4-5}
     & \textbf{Rec.} & \textbf{Prec.} & \textbf{Rec.} & \textbf{Prec.} \\
     \midrule
     \makecell[l]{GPT-3.5 (5)} & 75.4 $_{(0.6)}$ & 74.2 $_{(0.8)}$ & \textbf{80.4} $_{(0.3)}$ & \textbf{67.2} $_{(0.8)}$  \\
     \makecell[l]{GPT-3.5 (5-summ)} & 73.9 $_{(0.6)}$ & 72.4 $_{(0.3)}$ & 76.9 $_{(0.3)}$ & 59.4 $_{(0.4)}$  \\
     \makecell[l]{GPT-3.5 (5-snip)} & 60.5 $_{(0.3)}$ & 62.6 $_{(1.0)}$ & 68.1 $_{(1.4)}$ & 59.4 $_{(1.1)}$  \\
     \makecell[l]{GPT-3.5 (10)} & \textbf{75.8} $_{(0.6)}$ & \textbf{77.9} $_{(1.0)}$ & 78.6 $_{(0.8)}$ & 65.6 $_{(0.9)}$  \\
     \makecell[l]{GPT-4 (5)} & 69.3 $_{(0.8)}$ & 75.7 $_{(0.8)}$ & 76.0 $_{(0.3)}$ & 66.1 $_{(0.7)}$ \\
     \makecell[l]{LLaMA-3 (5)} & 56.9 $_{(1.0)}$ & 64.3 $_{(0.4)}$ & 60.3 $_{(0.4)}$ & 57.9 $_{(1.2)}$ \\
     \makecell[l]{LLaMA-3 (10)} & 57.7 $_{(1.0)}$ & 66.1 $_{(1.2)}$ & 58.2 $_{(1.5)}$ & 55.2 $_{(1.4)}$ \\
    \bottomrule
  \end{tabular}
  }
  \caption{Results on ASQA when only outputs containing positional fine-grained citations are evaluated. We omit the string "-psg" in the model settings for clarity. The best performances are highlighted in bold.}
  \label{table:alce_and_aliice_asqa}
\end{table}

\begin{table}\small
  \centering
  \resizebox{\linewidth}{!}{
  \renewcommand\arraystretch{1.3}
  \begin{tabular}{lcccc}
    \toprule
    \makecell[l]{\multirow{3}{*}{\textbf{Model ($k$-psg-form)}}}  & \multicolumn{2}{c}{\textbf{ALiiCE}} & \multicolumn{2}{c}{\textbf{ALCE}} \\
    \cmidrule(lr){2-3} \cmidrule(lr){4-5}
     & \textbf{Rec.} & \textbf{Prec.} & \textbf{Rec.} & \textbf{Prec.} \\
     \midrule
     \makecell[l]{GPT-3.5 (5)} & 40.1 $_{(1.8)}$ & \textbf{50.0 $_{(2.8)}$} & \textbf{48.1 $_{(2.4)}$} & \textbf{44.2 $_{(2.2)}$}  \\
     \makecell[l]{GPT-3.5 (5-summ)} & 35.9 $_{(2.5)}$ & 35.1 $_{(2.6)}$ & 42.9 $_{(1.2)}$ & 32.4 $_{(1.5)}$ \\
     \makecell[l]{GPT-3.5 (5-snip)} & 39.6 $_{(2.4)}$ & 39.1 $_{(1.3)}$ & 43.4 $_{(3.5)}$ & 34.5 $_{(2.7)}$  \\
     \makecell[l]{GPT-3.5 (10)} & \textbf{44.2 $_{(0.7)}$} & 48.1 $_{(1.0)}$ & 46.7 $_{(0.5)}$ & 41.0 $_{(1.7)}$  \\
     \makecell[l]{GPT-4 (5)} & 40.5 $_{(1.9)}$ & 46.2 $_{(1.1)}$ & 44.6 $_{(3.6)}$ & 38.7 $_{(3.5)}$ \\
     \makecell[l]{LLaMA-3 (5)} & 41.1 $_{(1.6)}$ & 44.0 $_{(0.9)}$ & 43.4 $_{(1.5)}$ & 39.7 $_{(1.0)}$  \\
     \makecell[l]{LLaMA-3 (10)} & 41.8 $_{(1.7)}$ & 47.7 $_{(3.7)}$ & 43.6 $_{(3.0)}$ & 41.0 $_{(3.6)}$ \\
    \bottomrule
  \end{tabular}
  }
  \caption{Results on ELI5 when only outputs with positional fine-grained citations are evaluated. Other descriptions follow Table \ref{table:alce_and_aliice_asqa}.}
  \vspace{-10pt}
  \label{table:alce_and_aliice_eli5}
\end{table}

\subsection{Error Analyses}

We further analyzed the potential errors in ALiiCE, which mainly come from two aspects: 

\paragraph{Grammatical error} Grammatical errors in the sentence can lead to inaccurate parsing results. However, current LLMs exhibit strong grammatical capabilities ~\citep{zhao2023survey}, and after
our manual evaluation, the number of samples containing grammatical errors in LLMs' outputs is nearly zero, thus this type of error can be ignored.

\paragraph{Parsing error} Dependency tree parsing itself might contain errors. For example, in sentence "Other radiological signs of fetal death include gas in the fetus or in the portal and umbilical vessels [1], and Deuel's halo sign [2].", the atomic claim of citation "[2]" is parsed as "Other radiological signs of fetal death include gas Deuel 's halo sign" by SpaCy, which contains an extra word "gas" due to an error from dependency recognition.

Therefore, we conduct further experiment to test the potential impact of parsing errors on NLI. We firstly collect all the atomic claims from two datasets. Next, we utilize GPT-3.5 to refine each claim based on its original sentence (the prompt is provided in Appendix \ref{appendix:promtps}). And then we employ the NLI model to assess the entailment before and after the claim refinement. As indicated in Table \ref{table:error_analyses}, the result show that the proportion of claims with inconsistent NLI results is less than 1\% across both datasets. Therefore, the parsing error is unlikely to have a significant impact on the evaluation.

\section{Human Evaluation}

We conduct human evaluation to examine the correlation between ALiiCE and human judgment. Since ~\citet{gao-etal-2023-enabling, liu-etal-2023-evaluating} have thoroughly studied sentence-level citation evaluation, we only focus on LLMs' responses that include positional fine-grained citations. In addition to the citation recall and precision, we also consider: 1) the proportion of positional fine-grained responses to total responses, 2) the answer utility, which assesses whether the LLM's response is helpful in answering the question, and 3) the citation utility, evaluates whether the positional fine-grained citation is useful for the response. We recruit three annotators to evaluate the outputs of the models used in the previous experiments.

We observe that ALiiCE and human judgment show a strong correlation. The model rankings evaluated by ALiiCE align closely with those evaluated by human judgement. The average Cohen's kappa coefficients between ALiiCE and annotators for ASQA are 0.71 for citation recall and 0.62 for citation precision, demonstrating high consistency. In addition, responses containing fine-grained citations constitute a small proportion of the total output. For instance, the fine-grained output of GPT-3.5 (5-psg-summ) on ELI5 accounts for only 8\% of the total samples. This pattern is consistent with the results shown by CVCP. Details on the human evaluation are provided in the Appendix \ref{appendix:human_evaluation_details}.

\section{Discussion}
\label{sec:discussion}

\begin{figure}[t!]
  \includegraphics[width=\linewidth]{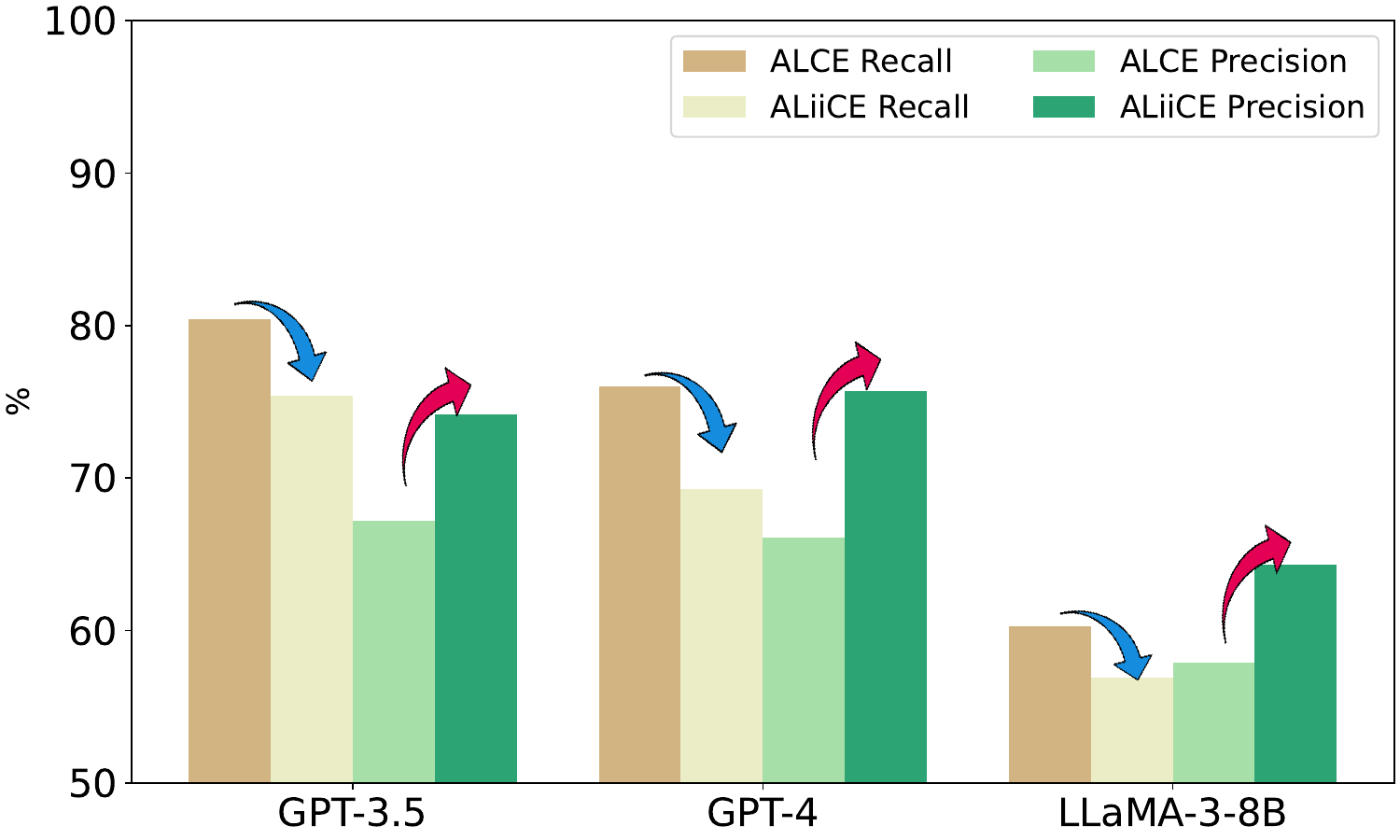}
  \caption {Comparison of citation recall and precision between ALCE and ALiiCE across three models using the 5-psg setting on ASQA. ALiiCE achieves lower citation recall and higher citation precision.}
  \label{fig:alce_and_aliice}
\end{figure}

Based on the experimental results and observations, we discuss our insights on the task of positional fine-grained citation generation, as follows:

\paragraph{ALiiCE has a higher decision threshold.} Compared to ALCE, ALiiCE calculates lower citation recall, but higher citation precision. And this difference becomes more dramatic when only positional fine-grained citation outputs are evaluated, as illustrated in Table \ref{table:alce_and_aliice_asqa} and Table \ref{table:alce_and_aliice_eli5}. We can observe this more intuitively in Figure \ref{fig:alce_and_aliice}. This means that ALiiCE has a higher decision threshold, indicating that ALiiCE is more conservative, only considering a citation correct when it has a high level of confidence. This is more beneficial for the citation generation task because the higher decision threshold encourages more accurate and relevant citations, reducing the likelihood of misleading information, which is particularly crucial in professional and high-risk fields (e.g., law and medicine) where incorrect citations can lead to serious consequences.

\paragraph{Open-source LLMs display great progress.} LLaMA-3 narrows the gap between open-source LLMs and closed-source LLMs in the citation text generation task. In previous studies, the citation quality of open-source LLMs is significantly worse than that of closed-source LLMs \citep{gao-etal-2023-enabling, huang2024training}. However, our experimental results show that the citation recall and precision of GPT-4 with 5-passages are only improved by 20.0\% and 14.6\%, respectively, compared to LLaMA-3-8B with 5-passages on ELI5. Additionally, LLaMA-3-8B has a higher CVCP and exhibits greater fluency, than both GPT-3.5 and GPT-4.

\paragraph{Rethinking citation quality through the lens of citation utility.} Our human evaluation indicates that citation utility and citation quality do not show a strong correlation. And our annotators observe that in some responses, even when the citation utility score is zero, the citation quality remains high. Thus, existing citation quality metrics can only evaluate the correctness of citation marks for each claim, but they fail to assess the utility of these marks, as being correct is not equal to being useful. We believe that this is worth further exploration in future research on citation evaluation methods.

\paragraph{How to study positional fine-grained citation generation?} Through our observation of the fine-grained responses, we find that most atomic claims with sufficient citation utility, exhibit certain logical relationships, such as parallelism, causality, and transitions. Under these logical structures, positional fine-grained citations often have better utility and significantly enhance user-friendliness. Constructing reasoning paths for multi-step retrieval and generation can establish clearer logical relationships for long-form responses, thereby promoting fine-grained citations. Additionally, in the supervised learning method, creating labeled data presents a significant challenge. ~\citet{ye2024effective} design an algorithm for automatically annotating citation marks at sentence-level. However, this method becomes more challenging for positional fine-grained citation generation. Similarly, we recommend constructing supervised labels by multi-hop QA datasets and also combining sentence-level citation sequences to ensure generalization.

\section{Related Work}

\paragraph{Attribution}
Attribution refers to the ability of LMs to generate and provide evidence ~\citep{li2023attributionsurvey}. 
The source of attribution can be pre-training data ~\citep{pretrainattribution1, pretrainattribution2}, or out-of-model knowledge ~\citep{ragattribution1, ragattribution2}. When the source is documents, citation is a common form of attribution \citep{kamalloo2023hagrid}. ~\citet{ye2024effective, huang2024training} study generating response and citations simultaneously, while \citet{posthoc1,posthoc2} research on adding citations in the post-hoc stage.

\paragraph{Retrieval-Augmented Generation}
Retrieval-augmented generation \citep[RAG;][]{rag} combines the strengths of information retrieval and generation models, demonstrating improvement in several NLP tasks. The primary methods for incorporating external knowledge into generation include modifying model parameters \citep{sen-etal-2023-knowledge} and Chain-of-Thought \citep[CoT;][]{cot, xu2024searchinthechain}. Since RAG exhibits a black-box nature \citep{gao2024retrievalaugmented}, adding citations in response can effectively mitigate the hallucination problem and enhance verifiability.

\paragraph{Citation Evaluation}
The current citation evaluation methods are mainly performed by human evaluation, which is costly and time-intensive \citep{chen2023understanding}. Thus, automatic evaluation methods are studied, including classification-based metrics \citep{liu-etal-2023-evaluating, yue-etal-2023-automatic} and quantitative metrics \citep{gao-etal-2023-enabling, ragattribution2}.
In specific domains, \citet{li-etal-2022-corwa, li2024relatedworkcitationtext} study the citation generation for academic writing.
However, most research is primarily sentence-level, leading to issues with atomicity of claims \citep{malaviya2024expertqa} and ambiguity \citep{liu-etal-2023-evaluating}. We propose ALiiCE, the first evaluation method for positional fine-grained citations.

\section{Conclusion}

In this study, we propose ALiiCE, the first evaluation method for positional fine-grained citation generation. Our approach incorporates an algorithm for parsing atomic claims based on dependency analysis, along with three metrics designed to assess the quality of positional fine-grained citations.

We evaluate several LLMs and observe that currently, LLMs lack strong capabilities for generating fine-grained citations. We demonstrate the need of designing dedicated method for positional fine-grained citation evaluation and the effectiveness of ALiiCE in addressing this need. We also discuss some useful conclusions: 1) the latest open-source LLMs narrow the gap between them and closed-source LLMs in citation generation; 2) current metrics for citation quality lack consideration of citation utility; 3) the logical relationships between atomic claims can be considered when designing methods for positional fine-grained citation generation. We hope that our work can inspire more research into this underexplored task.

\section*{Limitations}
In the implementation of our parsing method, we only employ SpaCy to construct dependency trees. Other dependency analysis methods with higher accuracy can improve our benchmark, which are not evaluated in our work. In addition, dependency analysis may be primarily applicable to mainstream languages such as English. Thus directly transferring ALiiCE to other languages might result in reduced evaluation accuracy.

In our experiments, we only utilize the open-domain long-form QA datasets. However, positional fine-grained citation generation is applicable to a broader range of scenarios, such as academic writing and summarization. Therefore, it is necessary to expand the data domain of the benchmark.

\section*{Ethics Statement}

The citation generation task aims to enhance the credibility of the generative model, assist users in verifying information, and mitigate the spread of misunderstandings or incorrect information. Additionally, it helps reduce ethical risks by clarifying responsibilities and respecting intellectual property rights. This research utilizes publicly available datasets sourced from widely recognized and reputable repositories. We have ensured that all datasets used in this study comply with relevant data usage and privacy policies.

\section*{Acknowledgments}

This work was supported by the National Key R\&D Program of China (2023YFC3303800).

\bibliography{custom}

\clearpage
\appendix

\section{Experimental Setup Details}
\label{appendix:experimental_setup_details}

\subsection{Datasets}

We utilize two datasets of long-form QA, and a detailed description of them is as follows:

\paragraph{ASQA} This is an open-domain long-form QA dataset for ambiguous factoid queries, collected from AmbigQA ~\citep{min-etal-2020-ambigqa}. Each query is annotated with long-form answers and multiple sub query-answer pairs that should be answerable by the long-form answers. We only use the development split of ASQA, which has 948 queries.

\paragraph{ELI5} This is a dataset for long-form QA, collected from subreddit "Explain Like I’m Five". First, its queries are complex enough to encourage paragraph-length responses. Second, each query requires reference to multiple knowledge sources. We only employ 1,000 examples collected randomly from its validation split.

\subsection{Models}

For \hytt{LLaMA-3-8B}, we set top\_p=0.95 for Nucleus Sampling ~\citep{nucleussampling}. And we set the sampling temperature to 0.5 for all models.



\begin{table*}
  \centering
  \small
  \resizebox{\linewidth}{!}{
  \renewcommand\arraystretch{1.1}
  \begin{tabular}{llccccccc}
    \toprule
    \multirow{3}{*}{\textbf{Datasets}} & \multirow{3}{*}{\textbf{Model ($k$-psg-form)}} & \multirow{3}{*}{\textbf{PF Sample}} & \multirow{3}{*}{\textbf{Answer Utility}} & \multirow{3}{*}{\textbf{PF Citation Utility}} & \multicolumn{2}{c}{\textbf{Human}} & \multicolumn{2}{c}{\textbf{ALiiCE}} \\
    \cmidrule(lr){6-7} \cmidrule(lr){8-9}
    & & & & & \textbf{Rec.} & \textbf{Prec.} & \textbf{Rec.} & \textbf{Prec.} \\
    \midrule
    \multirow{7}{*}{ASQA} & GPT-3.5 (5-psg) & 71 $_{(7.5\%)}$ & 3.4 $_{(0.58)}$ & 0.62 $_{(0.78)}$ & 77.0 & 75.2 & 75.4 & 74.2 \\
    & GPT-3.5 (5-psg-summ) & 161 $_{(17.0\%)}$ & 3.2 $_{(0.64)}$ & 0.50 $_{(0.73)}$ & 70.1 & 69.8 & 73.9 & 72.4 \\
    & GPT-3.5 (5-psg-snip) & 190 $_{(20.0\%)}$ & 3.5 $_{(0.57)}$ & 0.47 $_{(0.68)}$ & 63.9 & 62.4 & 60.5 & 62.6 \\
    & GPT-3.5 (10-psg) & 129 $_{(13.6\%)}$ & 3.2 $_{(0.47)}$ & 0.49 $_{(0.72)}$ & \textbf{77.8} & \textbf{78.8} & \textbf{75.8} & \textbf{77.9} \\
    & GPT-4 (5-psg) & 140 $_{(14.8\%)}$ & 3.7 $_{(0.50)}$ & 0.63 $_{(0.69)}$ & 70.5 & 75.5 & 69.3 & 75.7 \\
    & LLaMA-3-8B (5-psg) & 286 $_{(30.2\%)}$ & 3.0 $_{(0.49)}$ & 0.45 $_{(0.66)}$ & 58.1 & 66.7 & 56.9 & 64.3 \\
    & LLaMA-3-8B (10-psg) & 252 $_{(26.7\%)}$ & 2.7 $_{(0.44)}$ & 0.41 $_{(0.63)}$ & 58.9 & 67.2 & 57.7 & 66.1 \\
    \midrule
    \multirow{7}{*}{ELI5} & GPT-3.5 (5-psg) & 85 $_{(8.5\%)}$ & 3.1 $_{(0.57)}$ & 0.43 $_{(0.67)}$ & 41.4 & \textbf{49.4} & 40.1 & \textbf{50.0} \\
    & GPT-3.5 (5-psg-summ) & 80 $_{(8.0\%)}$ & 2.8 $_{(0.56)}$ & 0.46 $_{(0.63)}$ & 37.1 & 35.8 & 35.9 & 35.1 \\
    & GPT-3.5 (5-psg-snip) & 104 $_{(10.4\%)}$ & 2.9 $_{(0.42)}$ & 0.51 $_{(0.60)}$ & 40.9 & 39.7 & 39.6 & 39.1 \\
    & GPT-3.5 (10-psg) & 132 $_{(13.2\%)}$ & 2.7 $_{(0.48)}$ & 0.46 $_{(0.63)}$ & \textbf{42.7} & 46.6 & \textbf{44.2} & 48.1 \\
    & GPT-4 (5-psg) & 119 $_{(11.9\%)}$ & 3.3 $_{(0.55)}$ & 0.50 $_{(0.62)}$ & 41.0 & 45.7 & 40.5 & 46.2 \\
    & LLaMA-3-8B (5-psg) & 230 $_{(23.0\%)}$ & 2.8 $_{(0.51)}$ & 0.36 $_{(0.70)}$ & 40.4 & 46.0 & 41.1 & 44.0 \\
    & LLaMA-3-8B (10-psg) & 207 $_{(20.7\%)}$ & 2.4 $_{(0.66)}$ & 0.33 $_{(0.61)}$ & 39.6 & 47.2 & 41.8 & 47.7 \\
    \bottomrule
  \end{tabular}
  }
  \caption{Human evaluation results on ASQA and ELI5. The value in bracket of PF Sample is the percentage of responses containing positional fine-grained citation to the total responses. The value in bracket of Answer Utility and PF Citation Utility is the Fleiss' Kappa coefficient of three annotators. Every value of human evaluation metrics in the table is the average of the results from three annotators.}
  \label{table:human}
\end{table*}

\section{Human Evaluation Details}
\label{appendix:human_evaluation_details}

We conduct human evaluation to examine the correlation between ALiiCE and human judgment. We manually inspect only those samples containing positional fine-grained citations in the model output, as these are aligned with our task requirements. We focus on five metrics, as detailed below:

\paragraph{PF Sample} This represents positional fine-grained sample, which is the quantity of responses containing positional fine-grained citations.

\paragraph{Answer Utility} Whether LLM's response is helpful in answering the question. We employ a 1-5 Likert scale, corresponding to Strongly Disagree, Disagree, Neutral, Agree, and Strongly Agree.

\paragraph{PF Citation Utility} Whether the positional fine-grained citation in the response is useful. We employ a binary annotation rule. PF Citation Utility is 1 when all the following conditions are met: 1) the positions of the fine-grained citations are reasonable (i.e., each citation corresponds to a clear atomic claim); 2) the fine-grained citations improve readability and user verifiability, and reduce ambiguity compared to sentence-level citations (for details, see Section \ref{sec:introduction}); and 3) there is no excessive or redundant citation. Otherwise, PF Citation Utility is 0. For example, "Filming began in late May 2015[3], and the movie was released on March 25, 2016[3]." contains redundant citations and does not improve readability or user verifiability. Therefore, its PF Citation Utility is annotated as 0. 

\paragraph{Citation Recall} This is the human-calculated citation recall. The annotator extract atomic claims manually and judge whether the cited passages can entail the claim. Its calculation is consistent with the description in Section \ref{sec:positional_fine_grained_citation_recall}.

\paragraph{Citation Precision} This is the human-calculated citation precision. The annotator judge whether each citation is redundant. Its calculation is consistent with the description in Section \ref{sec:positional_fine_grained_citation_precision}.

We recruit three annotators who are highly familiar with NLP research and well-acquainted with our work. The results of ASQA and ELI5 are shown in Table \ref{table:human}. Our analysis of the results is as follows:

\paragraph{ALiiCE and human judgement show a strong correlation.} We observe that the ranking of models evaluated by ALiiCE is consistent with the ranking based on human judgment. Furthermore, we calculate the Cohen’s kappa coefficient between ALiiCE and each annotator's judgement, and the average result shows that the coefficient of citation recall is 0.71, and the coefficient of citation precision is 0.62, demonstrating high consistency. Additionally, under positional fine-grained citations, human judgment does not align with ALCE's evaluation. This underscores the necessity of positional fine-grained evaluation methods, which cannot be substituted by sentence-level evaluation methods.

\paragraph{Current LLMs generate limited outputs containing positional fine-grained citation.} This indicates that LLMs still face difficulties in providing positional fine-grained citations, which echoes the observation in Section \ref{sec:overall_performances}.

\paragraph{No direct relationship between citation utility and answer utility.} We suggests that in long-form QA, fine-grained citations often occur within supplementary explanations rather than in the core sentences of the answers. However, answer utility is mainly contributed by the core sentence. Hence, there is no strong correlation between them.

\paragraph{Citation utility should be given serious consideration.} This conclusion is consistent with that in Section \ref{sec:discussion}. Our findings indicate that PF citation utility and citation quality do not demonstrate a strong correlation. Our annotators observed that in some samples, even when the utility score is zero, the citation quality remains high. For example, the PF Citation Utility of "Filming began in late May 2015[3], and the movie was released on March 25, 2016[3]." is 0, but the citation recall and citation precision are all 1. Therefore, existing citation quality metrics can only evaluate the correctness of citation marks for each claim, but they fail to assess the utility of these marks.

\section{Parsing Algorithm Details}
\label{appendix:parsing_algorithm_details}

In section \ref{sec:algorithm}, we simplify the process of parsing algorithm. In practice, we consider more details when decomposing modified trees for different claims. The dependency type, represented by the edge values in the dependency tree (which can refer to Figure \ref{fig:tree_6}), is a crucial factor in dependency analysis. Thus we take dependency types into account when modifying the dependency tree. Table \ref{table:relation} shows some common dependency types, and a comprehensive explanation can be found in the official SpaCy documentation\footnote{\url{https://spacy.io/api/dependencyparser}}.

Specifically, when calculating the modified tree for node $i$ and traversing to node $j$ in iteration, if the LCA node is neither node $i$ nor node $j$, a more detailed discussion by situations is as follows:

\begin{itemize}
    \item If there is a subtree between $T_i$ and $T_j$ with a dependency relation of "cc" between its root node and the LCA node (we refer to this subtree $T_c$), then we discuss:
    \begin{itemize}
        \item If $T_i$ is before $T_j$, then we discuss: If the LCA node is the root node of the dependency tree and $T_i$ has a dependency relation of "prep" or "advcl" with the LCA node, then replace the root node of the dependency tree with $T_i$; else, then remove $T_j$ and $T_c$.
        \item If $T_i$ is after $T_j$, then we discuss: If the LCA node is the root node of the dependency tree and $T_i$ has a dependency relation of "prep" or "advcl" with the LCA node, then remove $T_j$ and $T_c$; else, then replace the root node of the dependency tree with $T_i$.
    \end{itemize}
    \item Else, then we discuss: If the LCA node is the root node of the dependency tree, then replace the root node of the dependency tree with $T_i$; else, then remove $T_j$ from $T$.
\end{itemize}

\begin{table}[h]
  \small
  \centering
  \resizebox{\linewidth}{!}{
  \renewcommand\arraystretch{1.3}
  \begin{tabular}{ccc}
    \toprule
     \textbf{Relation Type} & \textbf{Explanation} \\
     \midrule
     acomp & adjectival complement \\
     advcl & adverbial clause modifier \\
     amod & adjectival modifier \\
     cc & coordination \\
     conj & conjunct \\
     nmod & nominal modifier \\ 
     nsubj & nominal subject \\
     nsubjpass & passive nominal subject \\
     pobj & object of a preposition \\
     prep & prepositional modifier \\
     punct & punctuation \\
    \bottomrule
  \end{tabular}
  }
  \caption{Several common types of dependency relation.}
  \label{table:relation}
\end{table}

\section{Parsing Examples}
\label{appendix:parsing_examples}

\begin{figure}[h]
  \includegraphics[width=\linewidth]{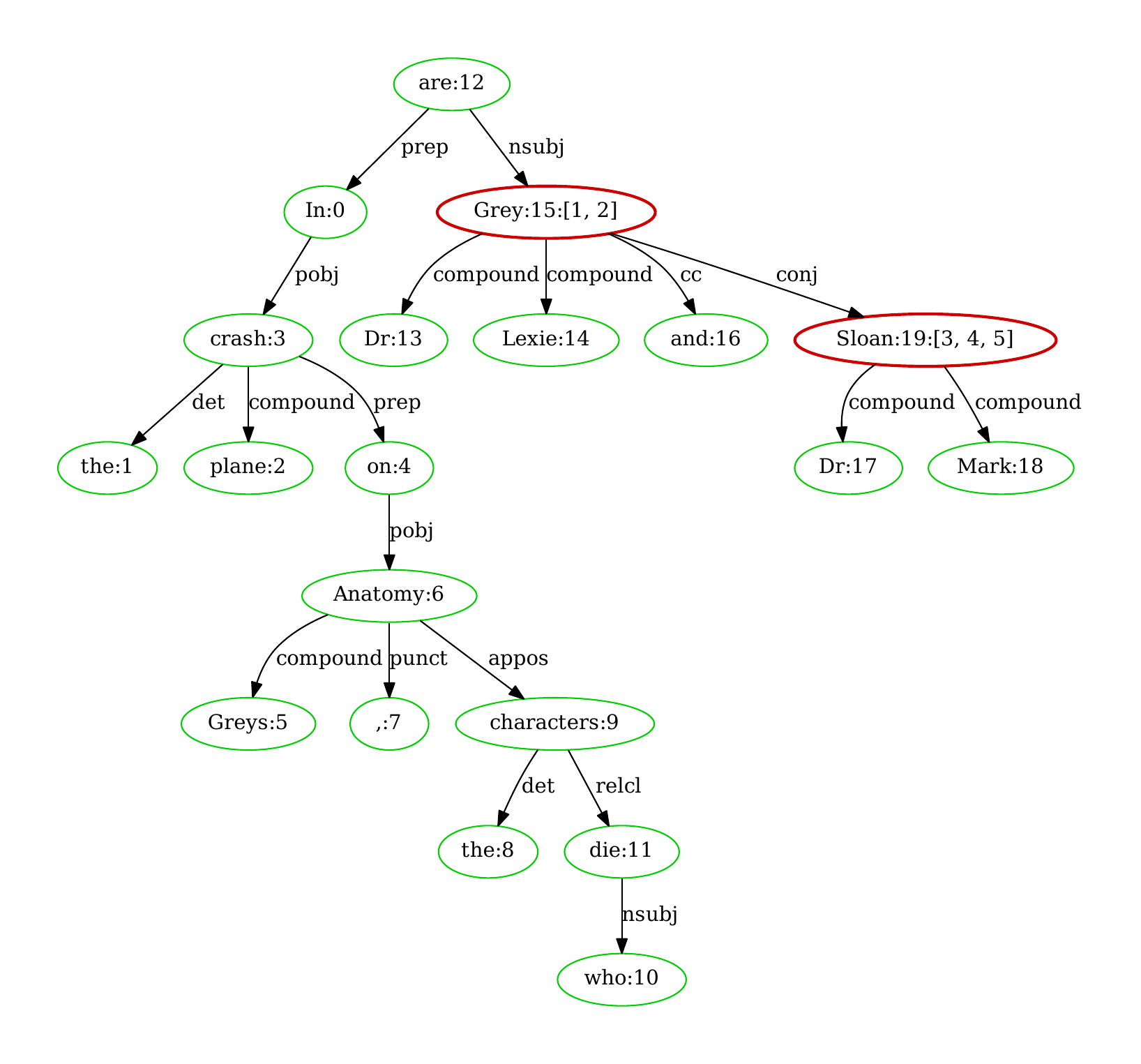}
  \caption {The dependency tree of sentence "In the plane crash on Grey's Anatomy, the characters who die are Dr. Lexie Grey [1][2] and Dr. Mark Sloan [3][4][5].", from the response generated by GPT-3.5 (5-psg). The query is "Who dies in the plane crash on greys?" from ASQA. The modified tree of claim corresponds to citation "[1][2]" is shown at Figure \ref{fig:tree_6_15}. The modified tree of claim corresponds to citation "[3][4][5]" is shown at Figure \ref{fig:tree_6_19}. }
  \label{fig:tree_6}
\end{figure}

\begin{figure}[h]
  \includegraphics[width=\linewidth]{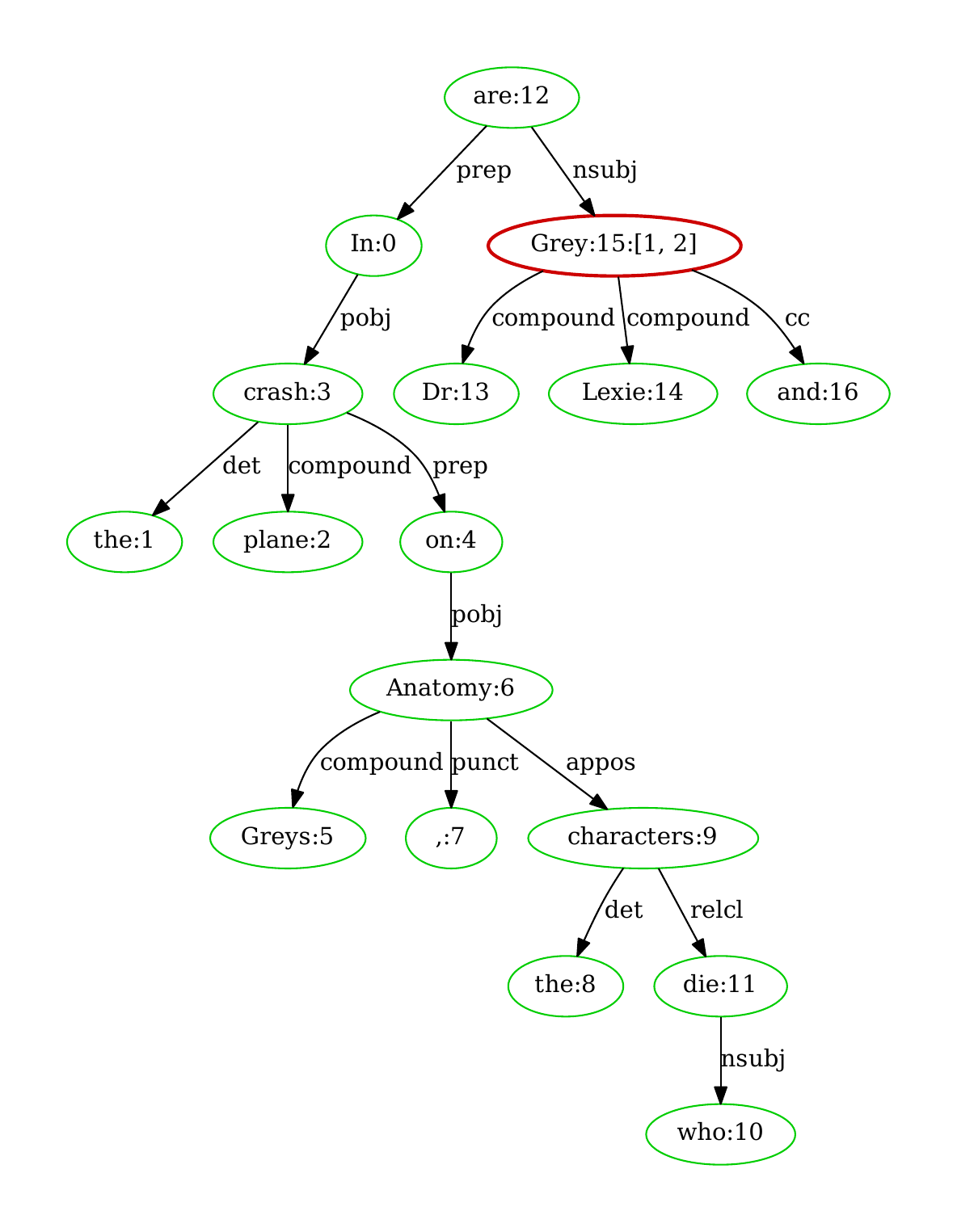}
  \caption {The modified tree of claim "In the plane crash on Greys Anatomy , the characters who die are Dr Lexie Grey and". This claim corresponds to citation "[1][2]" of sentence which is illustrated in Figure \ref{fig:tree_6}. }
  \label{fig:tree_6_15}
\end{figure}

\begin{figure}[h]
  \includegraphics[width=\linewidth]{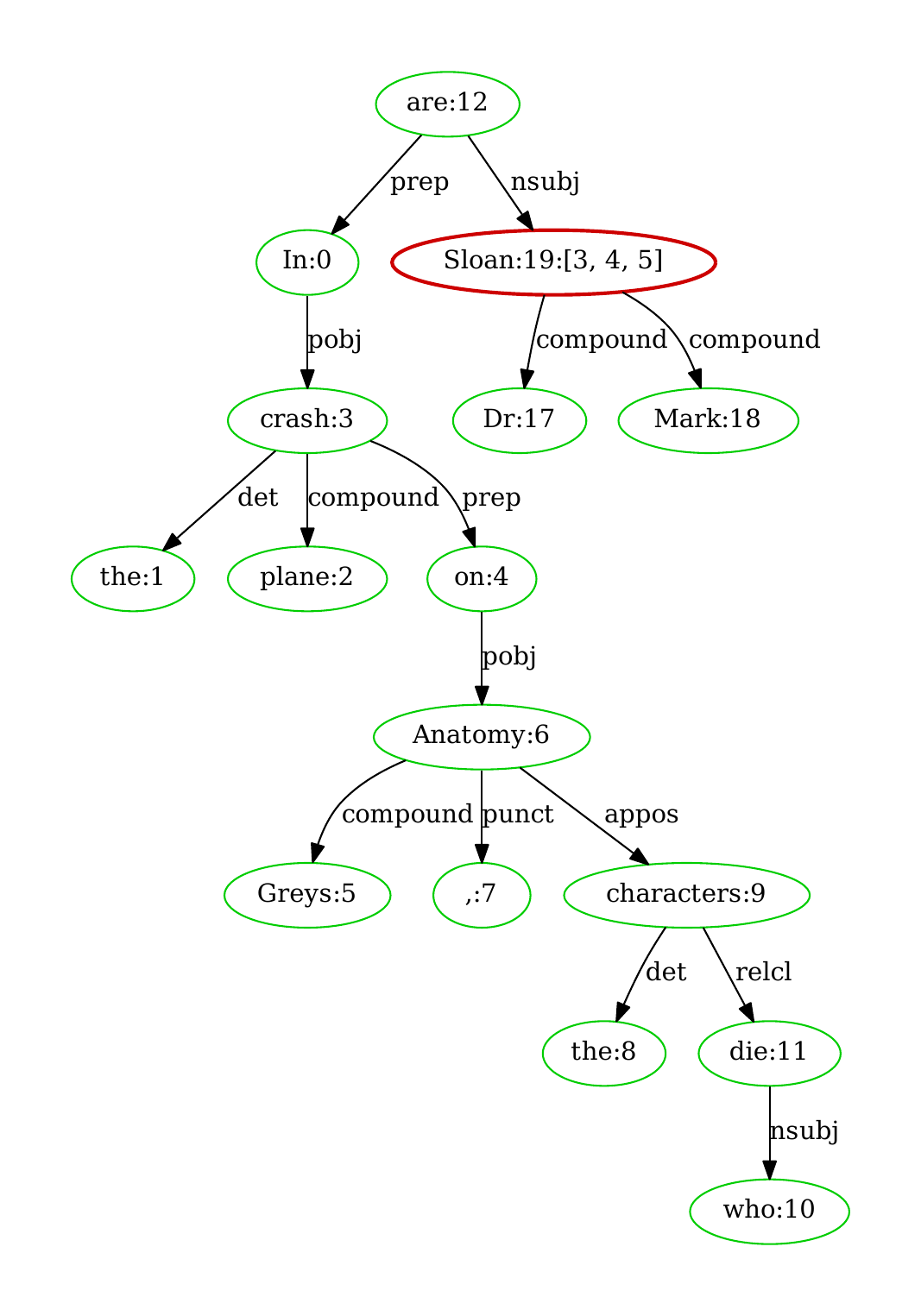}
  \caption {The modified tree of claim "In the plane crash on Greys Anatomy , the characters who die are Dr Mark Sloan". This claim corresponds to citation "[3][4][5]" of sentence which is illustrated in Figure \ref{fig:tree_6}.}
  \label{fig:tree_6_19}
\end{figure}

\begin{figure}[h]
  \includegraphics[width=\linewidth]{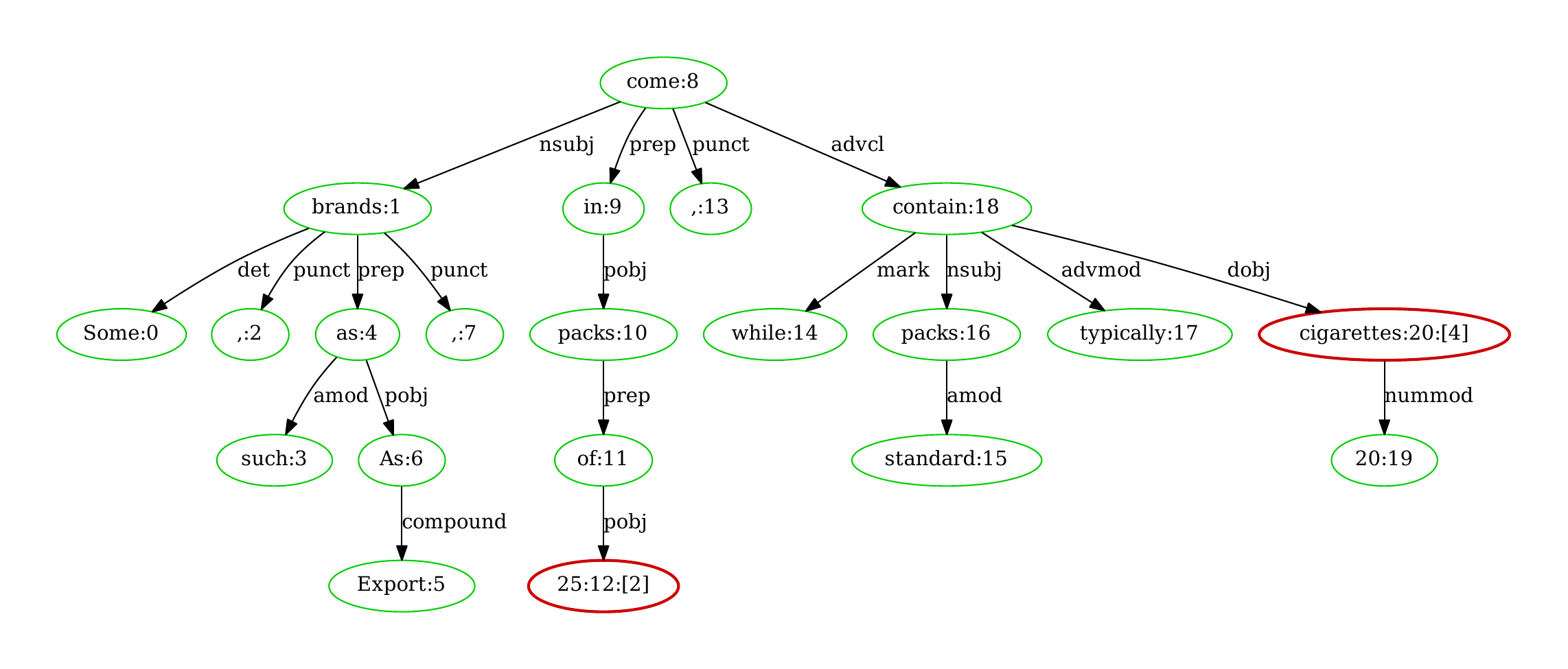}
  \caption {The dependency tree of sentence "Some brands, such as Export As, come in packs of 25 [2], while standard packs typically contain 20 cigarettes [4].", from the response generated by GPT-3.5 (5-psg). The query is "Number of cigarettes in a pack in usa?" from ASQA. The modified tree of claim corresponds to citation "[2]" is shown at Figure \ref{fig:tree_7_12}. The modified tree of claim corresponds to citation "[4]" is shown at Figure \ref{fig:tree_7_20}.}
  \label{fig:tree_7}
\end{figure}

\begin{figure}[h]
  \includegraphics[width=\linewidth]{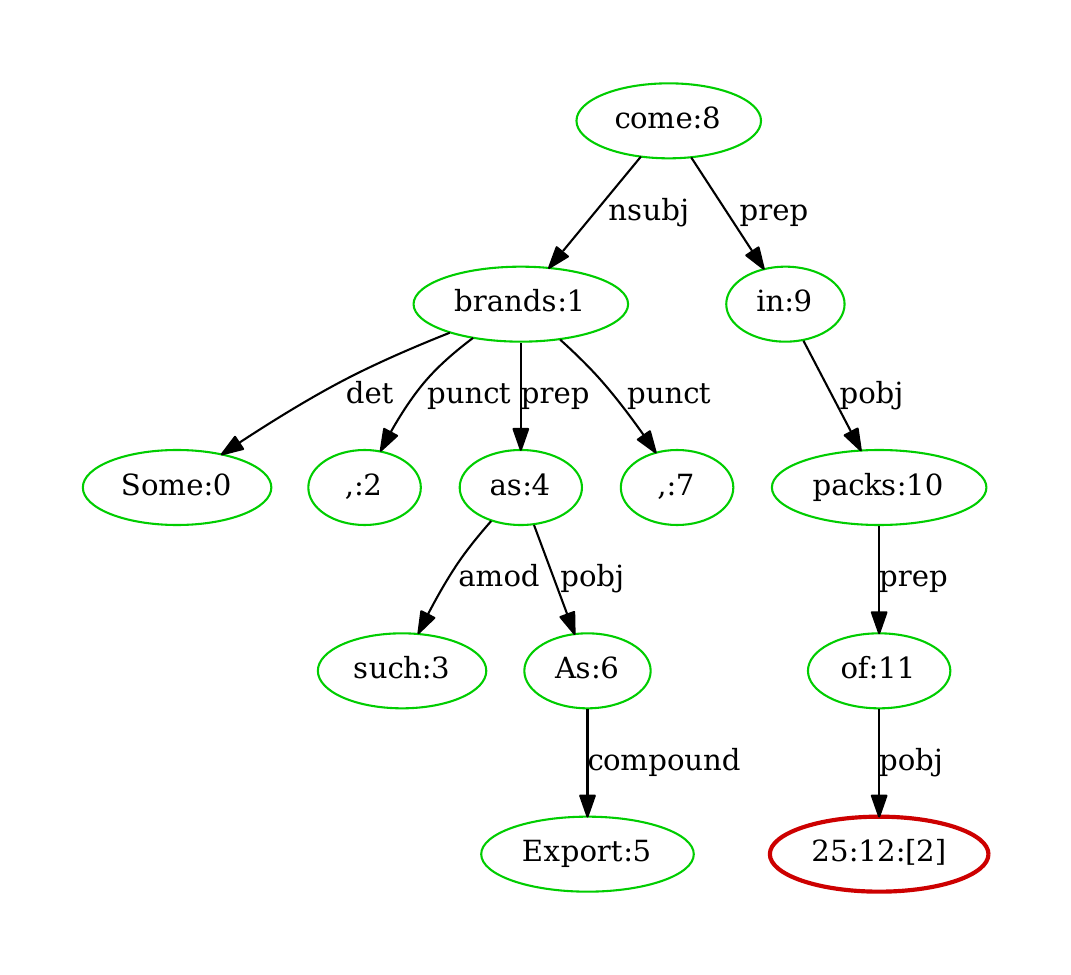}
  \caption {The modified tree of claim "Some brands , such as Export As , come in packs of 25". This claim corresponds to citation "[2]" of sentence which is illustrated in Figure \ref{fig:tree_7}.}
  \label{fig:tree_7_12}
\end{figure}

\begin{figure}[h]
  \includegraphics[width=\linewidth]{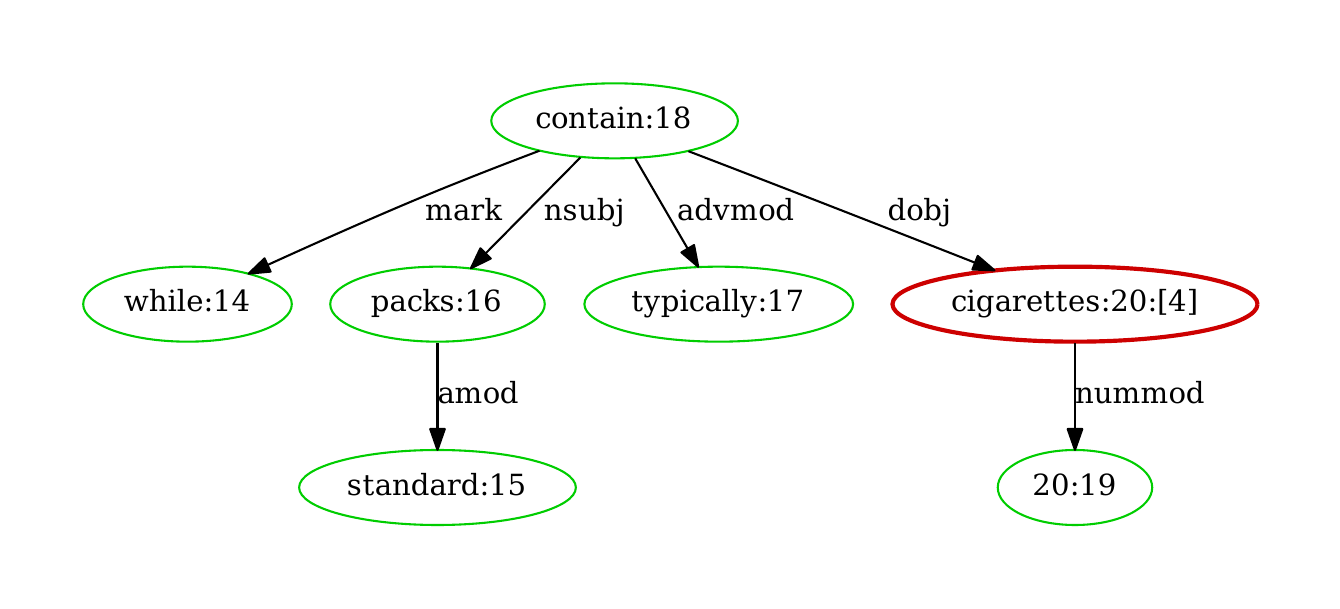}
  \caption {The modified tree of claim "while standard packs typically contain 20 cigarettes". This claim corresponds to citation "[4]" of sentence which is illustrated in Figure \ref{fig:tree_7}.}
  \label{fig:tree_7_20}
\end{figure}

\begin{figure}[h]
  \includegraphics[width=\linewidth]{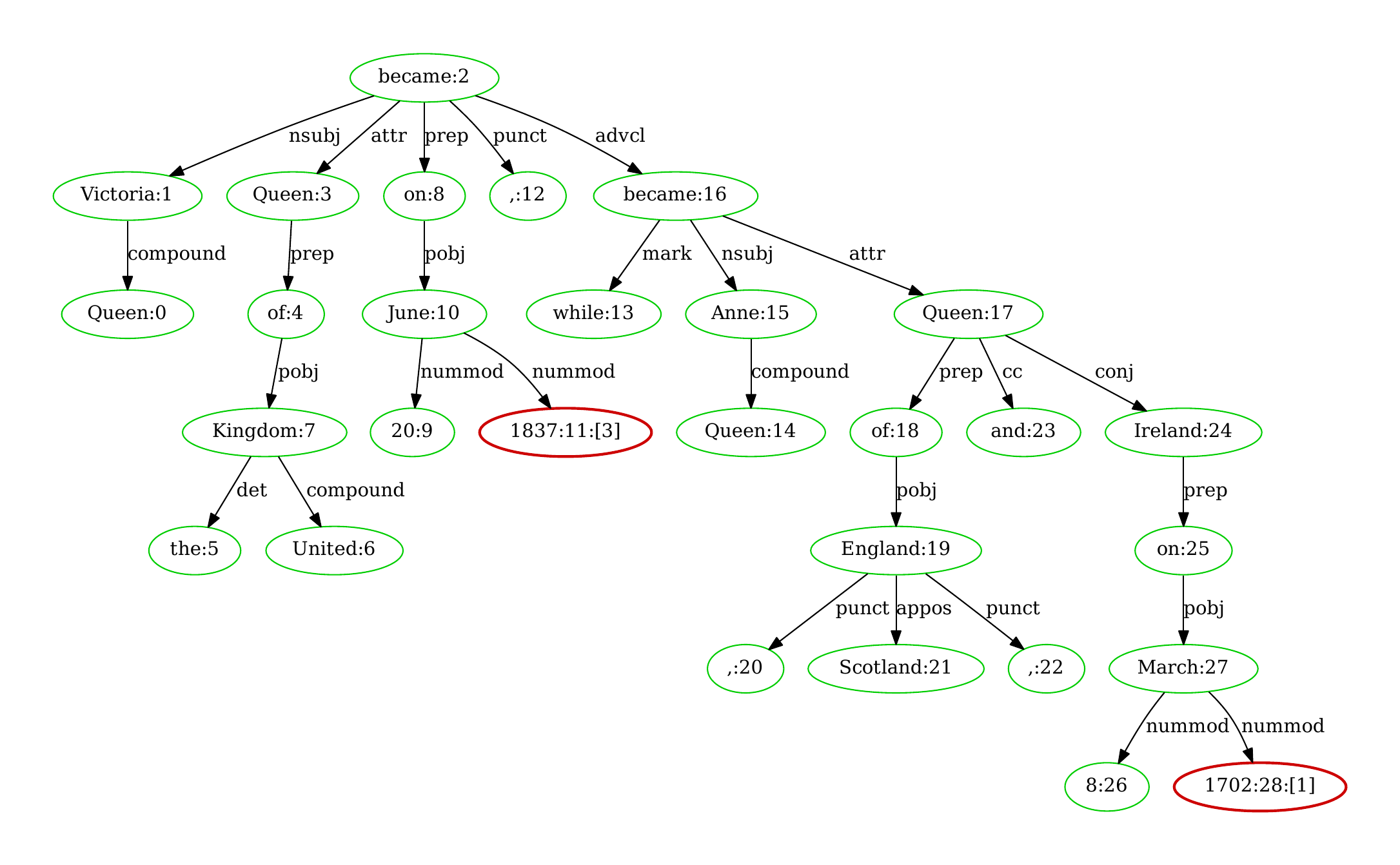}
  \caption {The dependency tree of sentence "Queen Victoria became Queen of the United Kingdom on 20 June 1837[3], while Queen Anne became Queen of England, Scotland, and Ireland on 8 March 1702[1].", from the response generated by GPT-3.5 (5-psg). The query is "When did the queen became queen of england?" from ASQA. The modified tree of claim corresponds to citation "[3]" is shown at Figure \ref{fig:tree_3_11}. The modified tree of claim corresponds to citation "[1]" is shown at Figure \ref{fig:tree_3_28}.}
  \label{fig:tree_3}
\end{figure}

\begin{figure}[h]
  \includegraphics[width=\linewidth]{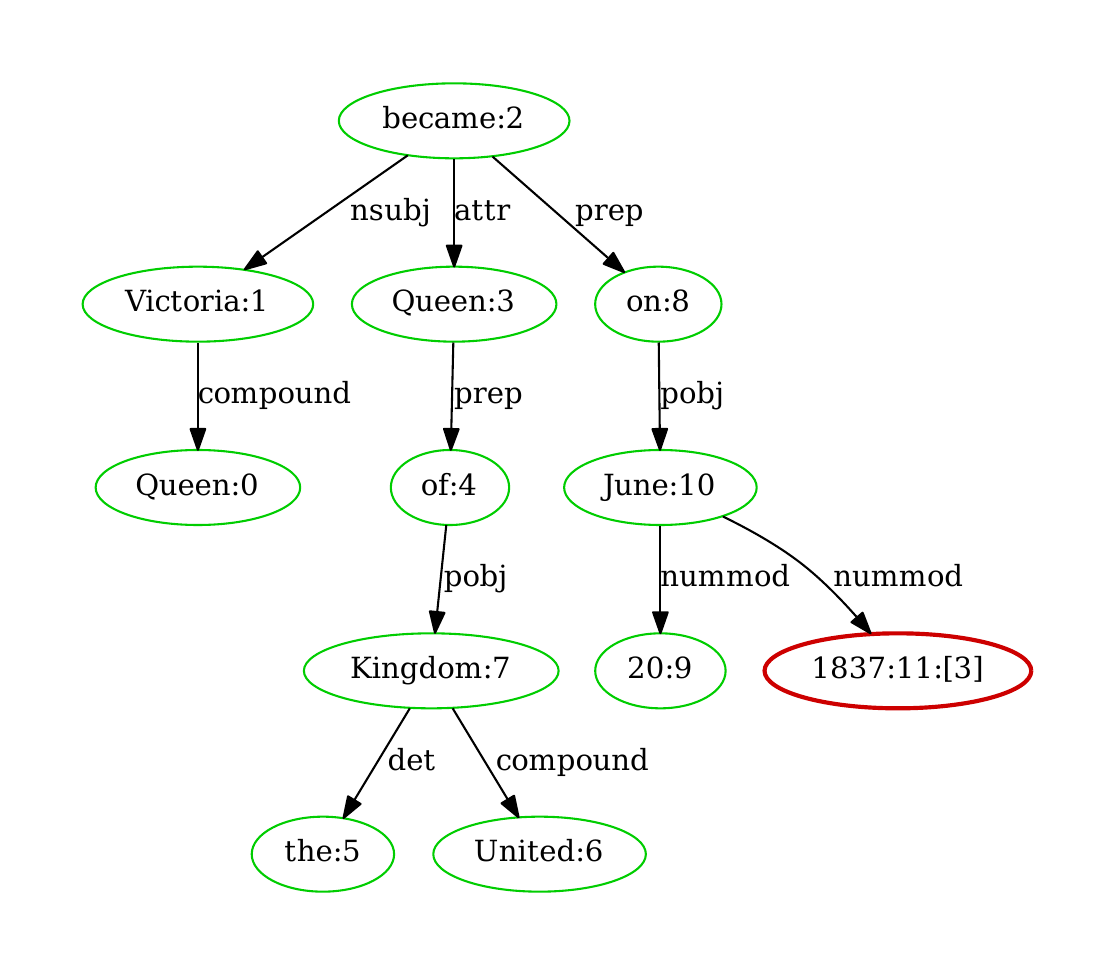}
  \caption {The modified tree of claim "Queen Victoria became Queen of the United Kingdom on 20 June 1837". This claim corresponds to citation "[3]" of sentence which is illustrated in Figure \ref{fig:tree_3}.}
  \label{fig:tree_3_11}
\end{figure}

\begin{figure}[h]
  \includegraphics[width=\linewidth]{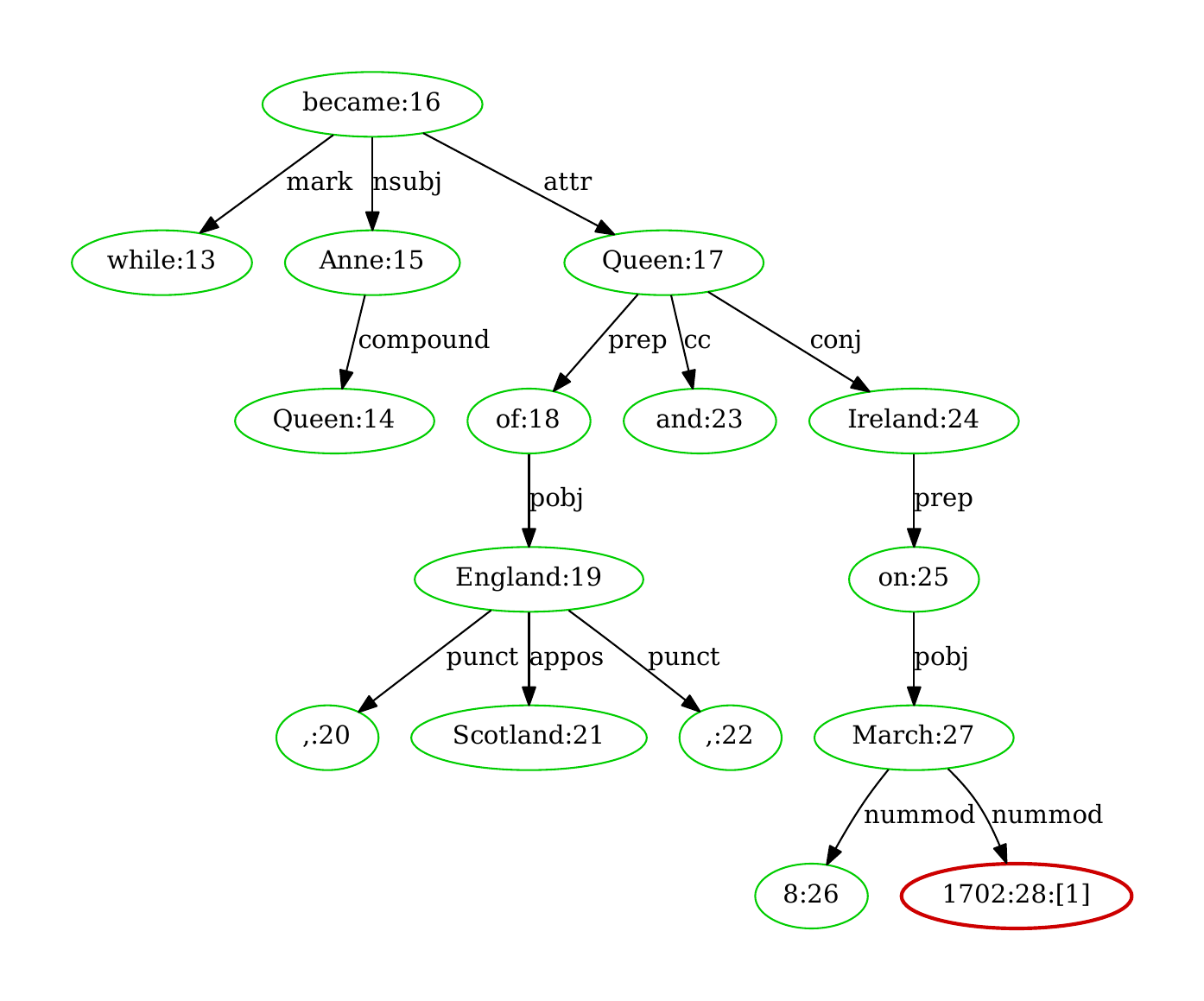}
  \caption {The modified tree of claim "while Queen Anne became Queen of England , Scotland , and Ireland on 8 March 1702". This claim corresponds to citation "[1]" of sentence which is illustrated in Figure \ref{fig:tree_3}.}
  \label{fig:tree_3_28}
\end{figure}

To improve the intuitiveness of the parsing algorithm, we present three straightforward examples (Figures \ref{fig:tree_6} to \ref{fig:tree_3_28}). Each figure shows a dependency tree, where each node represents a word node. For word nodes matched with citations (marked in red), the format of the node value is "word : index : citation marks", where "index" denotes the position of the word in the original sentence. For word nodes without citations (marked in green), the format of the node value is "word : index". The sentences to be parsed are all from the outputs of the GPT-3.5 (5-psg) on the ASQA dataset. 

Specifically, Figure \ref{fig:tree_6} illustrates the dependency tree for "In the plane crash on Grey's Anatomy, the characters who die are Dr. Lexie Grey [1][2] and Dr. Mark Sloan [3][4][5].", and Figures \ref{fig:tree_6_15} and \ref{fig:tree_6_19} display the modified trees for the two atomic claims in the output. Similarly, Figure \ref{fig:tree_7}, \ref{fig:tree_7_12}, and \ref{fig:tree_7_20} correspond to output "Some brands, such as Export As, come in packs of 25 [2], while standard packs typically contain 20 cigarettes [4].", and Figure \ref{fig:tree_3}, \ref{fig:tree_3_11}, and \ref{fig:tree_3_28} correspond to output "Queen Victoria became Queen of the United Kingdom on 20 June 1837[3], while Queen Anne became Queen of England, Scotland, and Ireland on 8 March 1702[1].". Notably, in the dependency tree shown in Figure \ref{fig:tree_6}, the LCA node of the two citation nodes is one of them. This structure represents the parallel relationship between two claims, which is a common form in positional fine-grained citations.

\section{Prompts}
\label{appendix:promtps}

We provide the prompts used in our experiments. We utilize the same prompt in fine-grained citation generation for all models, as shown in Table \ref{table:generating}. And Table \ref{table:rewriting} shows the prompt for claim rewriting employed in our error analysis experiments.

\section{CVCP Details}

In Appendix \ref{appendix:human_evaluation_details}, we preliminarily verify the consistency of CVCP with the degree of positional fine-grained citations. In this section, we further analyze the meaning of the CVCP value and provide a reference. We use the output of GPT-3.5, GPT-4, and LLaMA-3-8B with 5-psg generated from ASQA. We randomly select 200 responses containing fine-grained citations (denoted as $E$) and 200 responses without fine-grained citations (denoted as $F$). The CVCP for $E$ is 0.85, while for $F$ it was 0. We then randomly select 100 samples from each of $E$ and $F$ to form $G$, and repeat the calculation five times, resulting that the average CVCP of $G$ is 0.67. The reference of CVCP here is not entirely sufficient, as it would be more reasonable to use gold answers written by human experts. Thus, it is necessary to design a dedicated datasets for long-form QA with positional fine-grained citations, which should be addressed in future work.



\newpage
\clearpage

\begin{table*}[ht]
\centering
\small
\begin{tabular}{>{\raggedright\arraybackslash\tt}p{\linewidth}<{}}
\toprule
Instruction: Please provide an accurate and concise answer that includes fine-grained in-text citations immediately following the relevant information. Place the citation numbers within brackets directly after the facts they support.\\ \\

Citation format examples:\\
1. One of the most important areas is the automatic detection of vandalism[1][3] and data quality assessment in Wikipedia[2][4]. \\
2. Cups can be made of glass[1] or plastic[2][3].Wikipedia's community has been described as cultlike[1], although not always with entirely negative connotations[2]. \\
3. Wikipedia's community has been described as cultlike[1], although not always with entirely negative connotations[2]. \\ \\
            
Question: Who gets fired on grey's anatomy season 6? \\ \\

Documents [1] (Title: Now or Never (Grey's Anatomy)) an accident during the episode and dies in the season 6 premier. In the episode Cristina Yang (Sandra Oh), Alex Karev (Justin Chambers), George O'Malley (T.R. Knight), and Meredith Grey (Ellen Pompeo) are all sleeping and waiting for Izzie Stevens (Katherine Heigl) to wake up after the surgery. Derek Shepherd (Patrick Dempsey) comes up with an alternative treatment plan for Izzie, Miranda Bailey (Chandra Wilson) confronts Chief's Richard Webber (James Pickens Jr.) and Arizona Robbins (Jessica Capshaw), about the peds fellowship program. Yang deals with her relationship with Owen Hunt (Kevin McKidd) who helps George with career advice. The episode\\ \\

Documents [2] (Title: Grey's Anatomy) the head of neurosurgery and Meredith's love interest; Preston Burke (Isaiah Washington), the head of cardio, who becomes Yang's fiancé; and Richard Webber (James Pickens, Jr.), the Chief of Surgery and attending general surgeon, and the previous lover of Ellis Grey. In the sixth season, these residents are joined by Jackson Avery (Jesse Williams) and April Kepner (Sarah Drew), former Mercy-West residents who join Seattle Grace following an administrative merger. During the first six seasons, Burke, O'Malley, and Stevens all depart the series. In addition to Webber, Burke, and Shepherd, the surgical wing is primarily supervised by Addison Montgomery (Kate\\ \\

Documents [3] (Title: Grey's Anatomy (season 6)) Grey's Anatomy (season 6) The sixth season of the American television medical drama "Grey's Anatomy\", commenced airing on the American Broadcasting Company (ABC) in the United States on September 24, 2009, and concluded on May 20, 2010. The season was produced by ABC Studios, in association with Shondaland Production Company and The Mark Gordon Company; the showrunner being Shonda Rhimes. Actors Ellen Pompeo, Sandra Oh, Katherine Heigl, and Justin Chambers reprised their roles as surgical residents Meredith Grey, Cristina Yang, Izzie Stevens, and Alex Karev, respectively. Heigl was released from her contract in the middle of the season, while T.R.\\ \\

... \\ \\ 

Answer: {\color{blue} In "Grey's Anatomy" Season 6, the characters who get fired include Preston Burke, the head of cardio[2], and Izzie Stevens, portrayed by Katherine Heigl, who was released from her contract in the middle of the season[3]. Additionally, during the first six seasons, Burke, George O'Malley, and Izzie Stevens all depart the series[2].} \\
\bottomrule
\end{tabular}
\caption{The prompt used to generate a response. The blue text indicates the output of GPT-3.5. The question is from ASQA and the documents is retrieved from Wikipedia corpus by GTR.}
\vspace{-15pt}
\label{table:generating}
\end{table*}

\begin{table*}[ht]
\centering
\small
\begin{tabular}{>{\raggedright\arraybackslash\tt}p{\linewidth}<{}}
\toprule
Instruction: The following sentence may have some grammatical errors and may have some redundant ingredients. As long as it ensures fluency, you can delete some parts of the sentence that you think don't make sense.\\ \\
            
Original sentence: Other radiological signs of fetal death include gas in the fetus or in the portal and umbilical vessels, and Deuel's halo sign. \\ \\

Sentence to modify: Other radiological signs of fetal death include gas Deuel 's halo sign \\ \\

Modified sentence: {\color{blue} Other radiological signs of fetal death include Deuel's halo sign.} \\
\bottomrule
\end{tabular}
\caption{The prompt used to refine a claim. The blue text indicates the output of GPT-3.5. }
\vspace{-15pt}
\label{table:rewriting}
\end{table*}

\end{document}